\documentclass[lettersize,journal]{IEEEtran}
\usepackage{amsmath,amsfonts}
\usepackage{algorithmic}
\usepackage{algorithm}
\usepackage{array}
\usepackage[caption=false,font=normalsize,labelfont=sf,textfont=sf]{subfig}
\usepackage{textcomp}
\usepackage{stfloats}
\usepackage{url}
\usepackage{verbatim}
\usepackage{graphicx}
\usepackage{cite}
\usepackage{hyperref}
\usepackage{colortbl}
\usepackage{multirow}
\usepackage{booktabs} % 提供更好的表格线
\usepackage[table]{xcolor}
\begin{document}

\title{ILNet: Trajectory Prediction with Inverse Learning Attention for Enhancing Intention Capture}

\author{Mingjin Zeng, Nan Ouyang, Wenkang Wan, Lei Ao, Qing Cai, Kai Sheng
	\thanks{
		Mingjin Zeng, Nan Ouyang, Wenkang Wan, Lei Ao, Qing Cai, Kai Sheng are with the Key Laboratory of Collaborative Intelligence Systems, Ministry of Education, Xidian University.
		
		Corresponding author: Kai Sheng (e-mail: kaisheng@xidian.edu.cn). 
	}
}


\maketitle

\begin{abstract}
Trajectory prediction for multi-agent interaction scenarios is a crucial challenge.  
Most advanced methods model agent interactions by efficiently factorized attention based on the temporal and agent axes.
However, this static and foward modeling lacks explicit interactive spatio-temporal coordination, capturing only obvious and immediate behavioral intentions.  
Alternatively, the modern trajectory prediction framework refines the successive predictions by a fixed-anchor selection strategy, which is difficult to adapt in different future environments. 
It is acknowledged that human drivers dynamically adjust initial driving decisions based on further assumptions about the intentions of surrounding vehicles. 
Motivated by human driving behaviors, this paper proposes ILNet, a multi-agent trajectory prediction method with Inverse Learning (IL) attention and Dynamic Anchor Selection (DAS) module. 
IL Attention employs an inverse learning paradigm to model interactions at neighboring moments, introducing proposed intentions to dynamically encode the spatio-temporal coordination of interactions, thereby enhancing the model's ability to capture complex interaction patterns.
Then, the learnable DAS module is proposed to extract multiple trajectory change keypoints as anchors in parallel with almost no increase in parameters. 
Experimental results show that the ILNet achieves state-of-the-art performance on the INTERACTION and Argoverse motion forecasting datasets. 
Particularly, in challenged interaction scenarios, ILNet achieves higher accuracy and more multimodal distributions of trajectories over fewer parameters.
Our codes are available at \href{https://github.com/mjZeng11/ILNet}{https://github.com/mjZeng11/ILNet}.
\end{abstract}


\section{Introduction}
\IEEEPARstart{T}{rajectory} prediction aims to predict the future motions of dynamic targets (e.g., vehicles and pedestrians) by analyzing historical data and environmental information.
In multi-agent interaction scenarios, complex behavioral changes and high uncertainty vastly increase prediction difficulty.
It requires models to deeply comprehend driving contexts, including static road structures, historical participant interactions, and behavioral intentions \cite{liang2020learning,mozaffari2020deep,deng2025social}. 

The prior works devise a variety of scene element representations, such as rasterized images \cite{chai2019multipath,hong2019rules}, vectors \cite{zeng2021lanercnn,gao2020vectornet}, and point clouds \cite{ye2021tpcn}. 
These elements are encoded by using CNN \cite{nikhil2018convolutional,mohamed2020social}, LSTM \cite{zhang2019sr}, and Transformers \cite{yu2020spatio} to predict agent future trajectories. 
However, such methods are easily affected by high-frequency modes, limiting accurate multimodal predictions.
Some methods introduce candidate endpoints \cite{zhao2021tnt,gu2021densetnt}, future distribution heatmaps \cite{gilles2022gohome}, and trajectory scoring \cite{chai2019multipath,phan2020covernet} to generate multimodal trajectories. 
Whereas, they need to select the appropriate trajectories from many candidates, causing inefficiency in complex traffic scenarios.
Several element factorized attention-based works  \cite{nayakanti2023wayformer,ngiam2021scene,zhou2022hivt} effectively improve the prediction accuracy, but are not conducive to online prediction. 
QCNet \cite{zhou2023query} introduces a query-centric encoding paradigm of elements that balances online inference time and prediction accuracy. 
Some two-stage prediction methods \cite{shi2024mtr++,tang2024hpnet,zhou2023qcnext,wang2023ganet,zhou2024smartrefine} enhance multimodal adaptability by refining the initial trajectory predictions with different strategies.

\begin{figure}[tp] % [h] 表示尽量在当前位置插入图片
	\centering
	\includegraphics[width=0.46\textwidth]{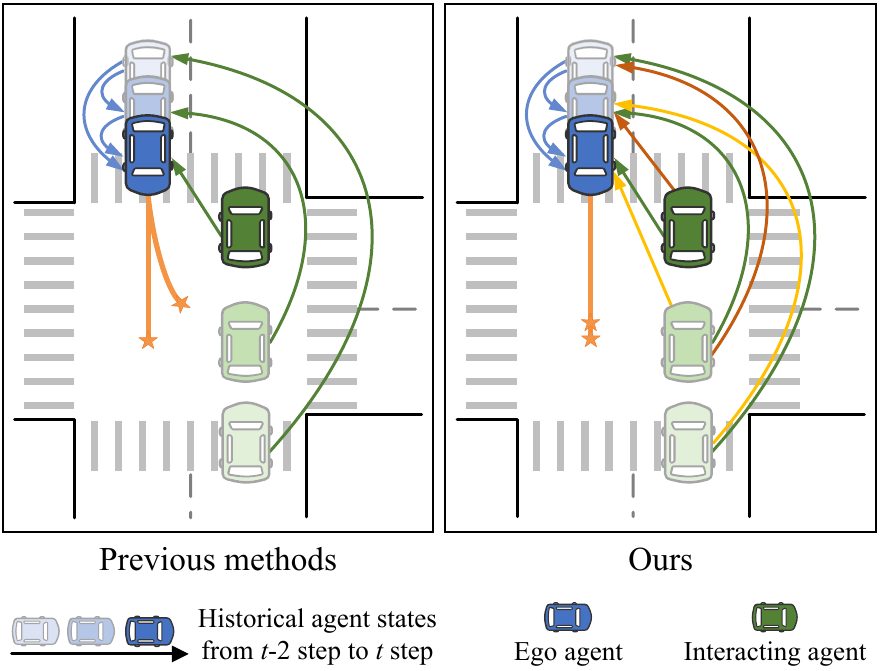} % 设置图片宽度为文本宽度的 70%
	\caption{ The difference between previous agent-agent interaction methods and ours.} % 设置图片标题
	\label{fig:fig1} % 设置图片标签，用于在文中引用
\end{figure}
The element interaction architectures mainly include agent-map interactions and agent-agent interactions. 
Previous advanced methods \cite{shi2024mtr++,ngiam2021scene,zhou2023query,tang2024hpnet} addressed the encoding of agent interactions using Temporal attention and Agent attention. 
The former independently models the temporal continuity of ego trajectories, while the latter models agent interactions independently of specific timestamps, as shown in Fig. \ref{fig:fig1}. 
However, this static and foward modeling lacks explicit interactive spatio-temporal coordination, which only captures obvious and immediate intentions.
Meanwhile, the methods that refine initial trajectories still neglect in-depth modeling of the interactions between historical trajectories, limiting the model's ability to refine and infer multimodal future trajectories.
In interactive driving scenarios, human drivers will adjust initial driving decisions based on further assumptions about the intentions of interacting vehicles \cite{sun2018probabilistic}. 
This process is based on experience-driven multiple dynamic inference.
As a result, more fully dynamic modeling of the interactive historical trajectory is critical to generating more accurate initial predictions.
HPNet \cite{tang2024hpnet} improves prediction accuracy by modeling the interactions between successive predicted trajectories. 
It takes the midpoints of these trajectories as anchors and refines predictions by integrating future contexts within a fixed radius of the anchor. 
Nevertheless, the fixed-anchor selection strategy is hardly competent in acquiring key future contexts in different scenarios,  thus affecting the refinement. 
SmartRefine \cite{zhou2024smartrefine} adaptively selects anchors based on the properties of each scene and then iteratively refines prediction. 
However, adopting this strategy will significantly increase the inference time for successive prediction refinements, especially in the case of multi-agent simultaneous inference, which is unfriendly to real-time trajectory prediction.

To tackle these problems, we propose a novel multi-agent trajectory prediction method, named ILNet.
Firstly, the obvious fact is that the intention of ego agent is susceptible to the behavioral styles of interacting agents \cite{deng2024eliminating}. 
Drawing inspiration from experienced human drivers, we explicitly argue that trajectory prediction models should employ retrospective reasoning about historical interactions, that is, an in-depth review of historical behavioral evolution through visible future states.
Therefore, we design an Inverse Learning (IL) attention mechanism.
Specifically, given the historical information of $T$ timestamps, for ego agents at the timestamp ${{t}_{i\in [0,T-1]}}$, the states of interacting agents in the timestamp $t+1$ are knowable. 
We consider the states of the timestamp $t+1$ as prior information, and introduce the modeling of interaction between ego agents at the current timestamp and interacting agents at the future timestamp to obtain prior embeddings, then inversely learns the behavioral changes of interacting agents from the history timestamp to the future timestamp, generating proposed intention that guides the current timestamp interaction, as shown in Fig. \ref{fig:fig1}.
%It forms a dynamic encoding of the spatio-temporal coordination between historical trajectory interactions, which enhances the model's ability to understand delicate intentions.
In contrast to forward modeling interaction methods, the inverse learning paradigm transforms prediction into a more sound process of logical inference. 
By conditioning on known future outcomes as guidance, the model learns to connect future outcomes to various underlying intentions, and to identify how these intentions are reflected in the nuances of past interactions. 
It forms the dynamic coordination of spatio-temporal interactions, which reduces the ambiguity of subtle intentions brought about by a single historical perspective, leading to more accurate and comprehensive modeling of subtle intentions.
Furthermore, to address the problem that successive prediction refinement with fixed anchors is difficult to adapt to the dynamic future context, we propose a parallel Dynamic Anchor Selection (DAS) module. 
Based on end-to-end training, this module aims to extract the multi-dimensional local correlation features of predicted trajectories, and then identify the keypoint of predicted trajectory change to be used as the refinement anchor. 
Our insight is that the subsequent trajectories of keypoints exhibit more uncertainty, and thus more information about the context around the keypoints needs to be fused to improve the stability of trajectories.
The DAS module improves multi-agent prediction performance with almost no increase in overall inference time and parameters. 
Our contributions can be summarized as follows: 

1) We design an Inverse Learning (IL) attention and propose ILNet, a trajectory prediction method which enhances the interaction of historical trajectories. 
It aims to dynamically encode the spatio-temporal coordination of interactions so that subtle interactive intentions can be understood more precisely, displaying better stability and flexibility in different scenarios.

2) A Dynamic Anchor Selection (DAS) module is proposed to extract trajectory change keypoints as anchors in parallel with almost no increase in overall infer time. 
The prediction accuracy can be effectively improved by integrating more information about the scene contexts around the keypoints.

3) The ILNet outperforms state-of-the-art methods on the INTERACTION and Argoverse motion forecasting benchmarks. 
Particularly, in challenged interaction scenarios, ILNet achieves higher accuracy and more multimodal distributions of trajectories over fewer parameters.

\section{RELATED WORK}
\subsection{Interaction Model}
In driving scenarios, agent interactions are complex and dynamic. 
Early machine learning-based approaches \cite{wissing2018trajectory,schlechtriemen2015will} use the “divide and conquer” strategy, combining perceptions and regressions for intention classification and trajectory prediction, which improve prediction performance but lack generalizability. 
Deep Learning-based Interaction Prediction methods show stronger advantages. 
Social-LSTM \cite{alahi2016social} introduces social pooling to capture the hidden states of agent interactions, but ignores spatial interactions between agents and roads. 
Chen et al. \cite{chen2022intention} design a dynamic attention to capture the spatio-temporal dependence of vehicular social interactions. 
Applying attention directly across all agent and temporal features in a standard Transformer is computationally costly and causes the identity symmetry challenge \cite{ngiam2021scene}.
Therefore, AgentFormer \cite{yuan2021agentformer} introduces a temporal encoder and agent-aware attention for addressing the loss of time and agent information.
Alternatively, several works \cite{varadarajan2022multipath++,nayakanti2023wayformer,feng2023macformer} employ self-attention and cross-attention to independently capture interactions among agents, lane routes, and temporal states.
The graph-based attention methods \cite{hu2022scenario,mo2022multi,liu2024laformer,salzmann2020trajectron++} provide flexible solutions for modeling social interactions. 
They are more effective in capturing the sparse topology of roads and interactive networks.
Social-wagdat \cite{li2020social} proposes spatio-temporal graph double-attention to generate abstract node attributes that represent agent interactions. 
EvolveGraph \cite{li2020evolvegraph} utilizes GRU to model the evolution of fully-connected interaction graphs.
Grin \cite{li2021grin} models uncertainty from intentions and relationships, respectively. 
Several methods \cite{zhou2023query,zhou2023qcnext,wang2023prophnet} model agent interactions using query-centered factorized attention.
HPNet \cite{tang2024hpnet} models the interactions between successive predicted trajectories.
DifTraj \cite{liu2024diftraj} diffuses future trajectories by combining destination-based intrinsic intentions and extrinsic interactions.

Distinguishing from the above forward modeling interaction methods (including the temporal consistency of predictions), the Inverse Learning paradigm transforms prediction into a more sound process of logical inference.
It enables the dynamic coordination of spatio-temporal interactions, reducing the ambiguity of subtle intentions caused by forward perspective, enabling more accurate and comprehensive modeling of subtle intentions.
\begin{figure*}[tp] % [h] 表示尽量在当前位置插入图片
	\centering
	\includegraphics[width=1\textwidth]{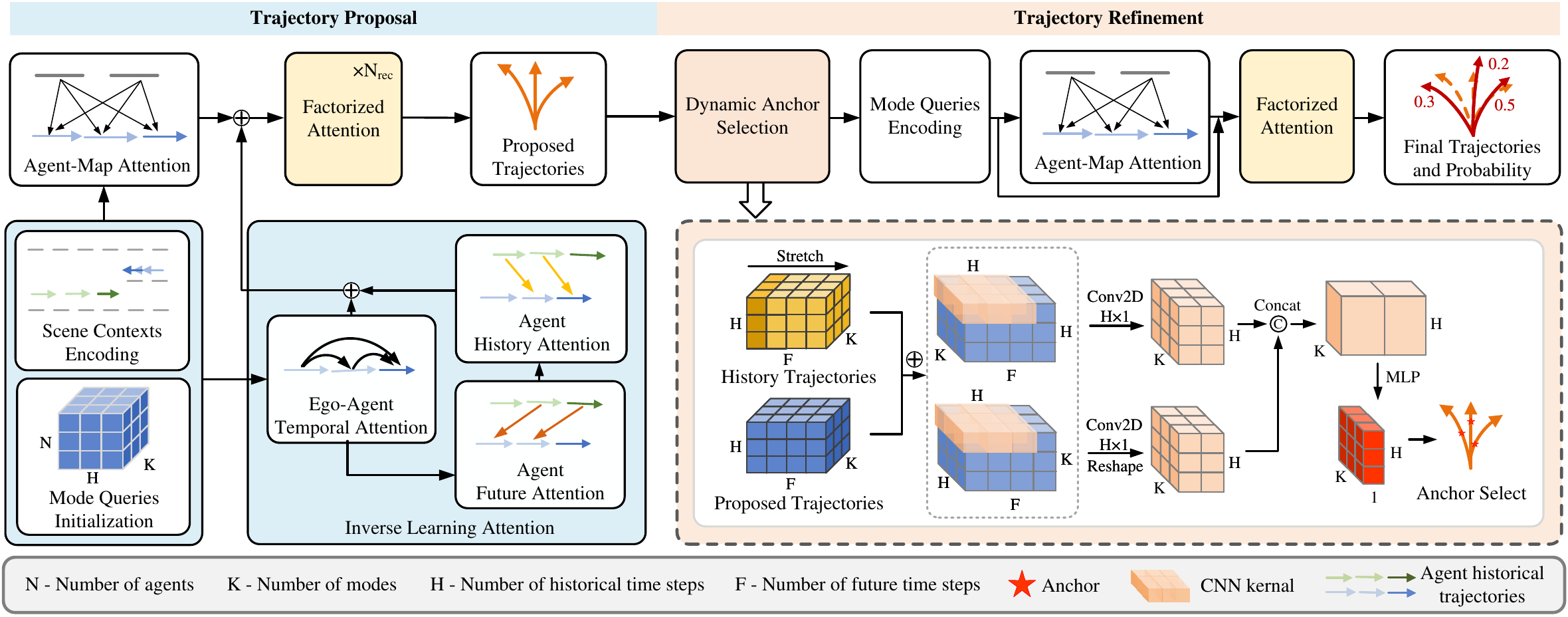} % 设置图片宽度为文本宽度的 70%
	\caption{An overview of ILNet. The proposed ILNet consists of Trajectory Proposal stage and Trajectory Refinement stage. The Inverse Learning Attention is utilized to obtain proposed intentions for dynamically encoding the spatio-temporal coordination of historical trajectory interactions. The Dynamic Anchor Selection module is applied to identify the keypoint of predicted trajectory change to be used as the refinement anchor.} % 设置图片标题
	\label{fig:fig2} % 设置图片标签，用于在文中引用
\end{figure*}
 \vspace{-1em} 
\subsection{Multimodal Output}
Multimodal prediction is another key task in trajectory prediction. 
Previous methods \cite{sadeghian2019sophie,salzmann2020trajectron++,yuan2021agentformer} employ generative models (e.g., GAN, VAE) to predict multiple trajectories. 
These methods require independent sampling and therefore do not guarantee diversity. 
Then, some methods introduce predefined trajectories \cite{chai2019multipath} and future distribution heat-maps \cite{gilles2022gohome}. 
mmTransformer \cite{liu2021multimodal} predicts different trajectory modes separately within predefined proposal areas. 
Predefined trajectories or potential targets are introduced to model the discrete distributions \cite{gu2021densetnt,zhao2021tnt,zhang2022trajectory}. 
MultiPath++ \cite{varadarajan2022multipath++} uses the latent anchor embeddings to capture different modes. 
ProphNet \cite{wang2023prophnet} introduces the usage of anchor-informed proposals to promote multimodal predictions. 
R-Pred \cite{choi2023r} designs a two-stage prediction method, including trajectory proposal and trajectory refinement. 
QCNet \cite{zhou2023query} proposes anchor-free proposal and anchor-based refinement with performance improvement. 
DiffusionDrive \cite{liao2025diffusiondrive} introduces a truncated diffusion strategy based on the anchored Gaussian distribution, achieving real-time and diverse trajectory generation.
Some recent works focus on anchor generation in the refinement stage, such as predicted trajectories \cite{shi2022motion,ye2023bootstrap}, high-confidence goals \cite{wang2023ganet}, trajectory midpoints \cite{tang2024hpnet}, and scenario-based selection \cite{zhou2024smartrefine}, all of which enhance the model's multimodal adaptability. 

Different from the above models, we propose a learnable end-to-end Dynamic Anchor Selection module to concurrently capture a key point of each trajectory change as the anchor. 
More accurate multimodal trajectories are generated by focusing on more important scene contexts using dynamic anchors.
\vspace{-1em} 
\section{Method}
\subsection{Problem Formulation}
Trajectory prediction is based on the $H$ history states of the given $N$ agents, denoted as ${{\{{{S}^{-h}},{{S}^{-h+1}},\ldots ,{{S}^{0}}\}}_{i\in [1,N]}}$, and high definition (HD) map information $M$ (e.g., road coordinates, topological connectivity,  and trafficability), to predict the future states of the $N$ agents, denoted as ${{\{{{S}^{1}},{{S}^{2}},\ldots ,{{S}^{F}}\}}_{i\in [1,N]}}$. 
The history states of the $i$-th agent include motion and attribute information. 
The future states ${{S}_{i,t}}\in{\mathbb{R}^{K\times F}}$ include the $F$ future positions $P({X}^{k,1\sim f},{Y}^{k,1\sim f})$ of $K$ different modal trajectories, denoting the predictions for the $i$-th agent at time $t$. 
We also assign a probability score $\pi$ to each modal trajectory, where the trajectory with the highest score represents the prediction most likely to match the ground truth path.
\subsection{Overview}
Our method consists of two stages, Trajectory Proposal and Trajectory Refinement, as shown in Fig. \ref{fig:fig2}. 
Trajectory Proposal: the scene contexts are encoded and the learnable mode queries are initialized.
Then, the Agent-Map attention and the Inverse Learning attention designed for enhancing interaction intention capture are conducted to obtain three kinds of embeddings, summing them to update the mode queries.
Next, the Factorized attention is performed on the mode queries, and repeated $N_{rec}$ times to enrich the expression of mode queries. 
The proposed trajectories are generated by decoding the mode queries through MLP. 
Trajectory Refinement: the key point of each proposed trajectory change is captured as the refinement anchor through the lightweight Dynamic Anchor Selection module.
The new anchor-based mode queries are encoded, and the Agent-Map attention and Factorized attention are executed once more, outputting the trajectory offset and probability score.
The final predicted trajectory is obtained by summing the proposed trajectory and trajectory offset.

\subsection{Trajectory proposal}
\subsubsection{Scene Context Encoding}
For each scene element (e.g., lanes and agents) encoding, we follow the query-centric encoding methods \cite{wang2023prophnet,zhou2023query,ye2023bootstrap}. 
Specifically, a local polar coordinate system is created for each element, its orientation is set to positive direction, and transform the global positions of surrounding elements into relative positions within this coordinate system. 

\textbf{Encoding Map Information. }A scene map contains $G$ lane segments ${{\{{{l}^{1}},\ldots ,{{l}^{g}}\}}_{g\in [1,G]}}$, each consisting of $O$ vectorized polylines ${\{{{p}^{1}},\ldots ,{{p}^{o}}\}}_{o\in [1,O]}$ (e.g., boundary lines and centerlines). 
The spatial states of each lane segment are represented by the position and orientation of the centerline midpoint.

Lane Segment-Polyline Interaction. The lane segments and polylines as nodes, nodes are featured by length embeddings. 
The mapping relationship between polyline and lane segment is represented by an edge, and its feature include $\{{{d}_{p}}_{l},{{\phi }_{p}}_{l},{{\varphi }_{p}}_{l},{{\gamma }_{p}}_{l}\}$. 
${{d}_{p}}_{l}$ denotes the distance from polyline to lane segment, ${{\phi }_{p}}_{l}$ denotes the orientation of edge in the lane segment local coordinate system, ${{\varphi }_{p}}_{l}$ denotes the relative orientation of polyline and lane segment, and ${{\gamma }_{p}}_{l}$ denotes the geometric attributes of polyline. 
Then, multi-layer perceptions (MLP) are used to obtain node embeddings ${{E}_{pl}}=MLP({{l}_{m}},{{p}_{m}})$ and edge embeddings $E_{pl}^{edg}=MLP({{d}_{p}}_{l},{{\phi }_{p}}_{l},{{\varphi }_{p}}_{l},{{\gamma }_{pl}})$, respectively. 
The lane segment local representations ${{l}_{m}}$ are integrated by fusing nodes and edges features via Graph Cross-Attention Network.

Lane Segment Interaction. The lane segments as nodes, node features as ${{l}_{m}}$. 
Edge features include$\{{{d}_{l}},{{\phi }_{l}},{{\varphi }_{l}},{{\gamma }_{l}}\}$, where ${{\gamma }_{l}}$ denotes the attributes of lane segments, including topological connectivity \cite{liang2020learning} (e.g., predecessor, successor, and neighbor) and multi-hop number.
We encode the edge features $E_{l}^{edg}=MLP({{d}_{l}},{{\phi }_{l}},{{\varphi }_{l}},{{\gamma }_{l}})$, and apply Graph Self-Attention Network to capture the topological associations between lanes and the interactions between distant lanes.
Finally the map embeddings ${{E}_{m}}\in \mathcal{{R}^{G\times D}}$ are generated, where $D$ denotes the feature dimension. 

\textbf{Encoding Agent Information}. We create the local coordinate system for each agent at each timestamp, again referencing the current position and orientation of agents. 
Non-location information of agents as node features ${{a}_{t}}=\{{{v}_{a}},{{s}_{a}},{{c}_{a}}\}$, where ${{v}_{a}}=({{d}_{a}}^{v},{{\phi }_{a}}^{v})$ is the velocity magnitude and velocity direction of agent, ${{s}_{a}}=({{l}_{a}},{{\omega }_{a}})$ is the agent shape, including length ${{l}_{a}}$ and width ${{w}_{a}}$, ${{c}_{a}}$ is agent category (e.g., vehicle and pedestrian). 
The node embeddings ${{E}_{a}}=MLP({{v}_{a}},{{s}_{a}},{{c}_{a}})$, where ${{E}_{a}}\in \mathbb{R}^{N\times H\times D}$. 
The edge features as $\{{{d}_{a}},{{\phi }_{a}},{{\varphi }_{a}},{{\lambda }_{a}}\}$, where ${{d}_{a}}$ is the length of edge, ${{\phi }_{a}}$ is the orientation of edge, ${{\varphi }_{a}}$ is the relative orientation of agents. 
${{\lambda }_{a}}$ is used only in the agent interactions involving time spans, representing the timestamp interval.  
We use MLP to generate edge embeddings $E_{a}^{edg}=MLP({{d}_{a}},{{\phi }_{a}},{{\varphi }_{a}},{{\lambda }_{a}})$ , where $E_{a}^{edg}\in \mathbb{R}^{Y \times D}$, $Y$ represents edge number.

%\vspace{1em} % 调整间距
\textbf{Agent-Map Attention.} The Agent-Map attention is designed for agents to extract the crucial environment constraint information. 
We first initialize the mode query embeddings ${{q}_{a}} \in{\mathbb{R}^{H\times K\times D}}$ for each agent, each embedding represents a possible future motion mode. 
They capture surrounding information based on the motion states of corresponding agents and continuously self-update in different interactions. 
Mode queries and lane segments are modeled as target nodes and source nodes, respectively. 
Within each timestamp, each target node will query its surrounding source nodes for a fixed spatial radius ${{R}_{m}}$, and then embed the contextual information of the lane segments through cross-attention. 
The process is described as:
\begin{equation}
	q_{sa} =\text{MHGCA}\left( q_{a}, \left[ E_{m} + E_{m}^{\text{edg}} \right], \left[ E_{m} + E_{m}^{\text{edg}} \right] \right)
\end{equation}
where ${{q}_{sa}}$ is spatial embedding, $\text{MHGCA}(q,k,v)$ represents the multi-head graph cross-attention with $q$ as query, $k$ as key, and $v$ as value. 
$[{{E}_{m}}+E_{m}^{edg}]$ is the sum of lane segment's node and edge features.

\subsubsection{Inverse Learning Attention}
The Inverse Learning (IL) attention is decomposed into three layers, including Ego-Agent Temporal attention, Agent Future attention, and Agent History attention. 
The Ego-Agent Temporal attention is responsible for modeling the trajectory evolution of ego agent over the historical timestamps, while the Agent Future attention and Agent History attention are novel modules introduced in our work. 
The Agent Future attention is used to obtain the prior influence of interacting agent on ego agent from the future timestamp.
The Agent History attention is utilized to inversely learn the motion changes of interacting agent between the future timestamp and the history timestamp. 
IL attention features spatio-temporal invariance and can be extended to realize stream scene encoding. 

\textbf{Ego-Agent Temporal Attention.} For each agent at timestamp $t$, the states of history timestamps with intervals $[0,t-1]$ can perform attention. 
First, we convert the global position of history timestamp into the relative position to obtain the edge features: the length and orientation of the edge, the relative orientation between the nodes, and the time interval. 
Then, each initial mode query performs the Ego-Agent Temporal attention for all source nodes at each timestamp:
\begin{equation}
	q_{ta} =\text{MHGCA}\left( q_{a}, \left[ E_{a} + E_{a}^{\text{edg}} \right], \left[ E_{a} + E_{a}^{\text{edg}} \right] \right)
\end{equation}
where ${{q}_{ta}}$ is temporal embedding. This attention ensures the model captures key historical information that is closely associated with the current timestamp state.

\textbf{Agent Future Attention. }The previous methods \cite{zhou2023qcnext,tang2024hpnet} perform trajectory interaction at the same timestamp after obtaining the embeddings ${{q}_{sa}}$ and ${{q}_{ta}}$. 
However, the Agent Future attention first focuses on the future timestamp states of interacting agents, modeling the interaction between ego agents at the current timestamp and interacting agents at the future timestamp to obtain prior embeddings. 
Edge indexes are constructed upon the connection relationships between the ego agents at timestamp $t\in [0,T-1]$ and the interacting agents at timestamp $t+1$. 
It is worth mentioning that these connection relationships can be reused in the next Agent History attention. 
Edge features include the length and orientation of the edge and the relative orientation between the nodes. 
For each temporal embedding ${{q}_{ta}}$ at each timestamp, queries the source nodes visible in future spacetime with the search radius ${{R}_{f}}$, and then performs the Agent Future attention to obtain the prior embeddings ${{q}_{fa}}$:
\begin{equation}
	q_{fa} = \text{MHGCA} \left( q_{ta}, \left[ E_{a} + E_{a}^{\text{edg}} \right], \left[ E_{a} + E_{a}^{\text{edg}} \right] \right)
\end{equation}

\textbf{Agent History Attention. }We set the span of Agent History attention from the timestamp $t-1$ to the timestamp $t+1$. 
On the one hand, this setting extends the windows of inverse learning in the temporal dimension, while not affecting subsequent agent interactions at the same moment, but rather combining them, maintaining spatio-temporal consistency in intention understanding. 
On the other hand, this setting allows the edges to be constructed by reversing the messaging direction of the upper attention. 
Then, for each prior embedding ${{q}_{fa}}$ at timestamp $t$ as a query, performing the Agent History attention on the interacting agents at timestamp $t-1$ within radius ${{R}_{h}}$ to obtain the inverse learning embeddings $q_{ia}$ as proposed intentions:
\begin{equation}
	q_{ia} = \text{MHGCA} \left( q_{fa}, \left[ E_{a} + E_{a}^{\text{edg}} \right], \left[ E_{a} + E_{a}^{\text{edg}} \right] \right)
\end{equation}
This inverse learning mechanism allows the model to infer past behavior from future perspectives, thus establishing tighter dynamic correlations between historical trajectories. 
It leads to better prediction performance compared to the forward inference patterns of obtaining proposed intentions, the quantitative result can be seen later in the ablation experiments.
The spatial embedding, temporal embedding, and inverse learning embedding are summed to obtain the multi-featured mode query embeddings ${{E}_{q}}$:
\begin{equation}
	E_{q} = q_{sa} + q_{ta} + q_{ia}
\end{equation}

\subsubsection{Factorized Attention}
Following the works \cite{tang2024hpnet,zhou2023query,zhou2023qcnext}, we serially execute the Factorized attention, including Agent attention, Historical Prediction attention, and Mode attention, modeling the interaction of mode queries in different dimensions. 
First, the Agent attention is used to extract the agent interaction at the same timestamp. 
Then, Historical Prediction attention and Mode attention are performed on the ego agent in the history dimension at the same mode and in the mode dimension at the same timestamp, respectively. 
The introduction of $q_{ia}$ allows the Agent Attention to model finer-grained social interactions by taking into account inferred intentions rather than just observed states. Further, this guides History and Modal Attention to achieve more precise alignment and judgments in modeling the past and evaluating the future.
The general form of each attention layer is as:
\begin{equation} 
	E_{q} = \text{MHGA} \left( E_{q}, \left[ E_{q} + E_{q}^{\text{edg}} \right], \left[ E_{q} + E_{q}^{\text{edg}} \right] \right)
\end{equation}
where $\text{MHGA}()$ is multi-head graph self-attention. The Factorized attention is repeated $N_{rec}$ times to enrich the expression of embedding $E_{q}$, and $E_{q}$ is decoded by MLP to generate the proposed trajectories $P_{i\in [1,N]}^{pro}\in {\mathbb{R}^{H\text{ }\!\!\times\!\!\text{ }K\text{ }\!\!\times\!\!\text{ }F\text{ }\!\!\times\!\!\text{ }2}}$.
%-------------------------------------------------------------------------
\subsection{Trajectory Refinement}
In order to reduce possible uncertainties and errors in the proposed trajectory, trajectory refinement reduces the risk of collision by considering the scenario interactions of each proposed trajectory, so that the final predicted trajectory is closer to the ground truth trajectory.

%\subsubsection{Dynamic Anchor Selection}
\textbf{Dynamic Anchor Selection. }First, we convert the historical trajectories $P_{i}^{his}\in {\mathbb{R}^{H\text{ }\!\!\times\!\!\text{ }K\text{ }\!\!\times\!\!\text{ }2}}$ and proposed trajectories  $P_{i}^{pro}$ into the scene-centeric representations in polar coordinates. 
Trajectory embeddings $E_{P,i}^{his}\in {\mathbb{R}^{K\text{ }\!\!\times\!\!\text{ }H}}$ and proposal embeddings $E_{p,i}^{pro}\in {\mathbb{R}^{K\text{ }\!\!\times\!\!\text{ }H\text{ }\!\!\times\!\!\text{ }F}}$ are obtained through different MLP layers. 
Then, $E_{P,i}^{his}$ is stretched along the last dimension, and $E_{P,i}^{{}}$ is obtained by summing with $E_{p,i}^{pro}$. 
Thus, each element on $E_{p,i}$ contains the spatial features of future timestamp and corresponding history timestamp. 
Next, we introduce 2D CNN blocks to extract local correlation features between ego future trajectories in different dimensions.
The process is denoted as:
\begin{equation}
	X_{i}^{H} = \text{Conv2D} \left( E_{p,i} \in \mathbb{R}^{H \times F \times K}, H, (H \times 1) \right)
\end{equation}
\begin{equation}
	X_{i}^{K} = \text{Conv2D} \left( E_{p,i} \in \mathbb{R}^{K \times F \times H}, K, (H \times 1) \right)
\end{equation}
where $u$ in $C\text{onv}2D(u,v,w)$ is the feature, $v$ is the Conv channel number, $w$ is the Conv kernel size. 
${\mathbb{R}^{c\times a\times b}}$ is the channel, length, and width of feature, respectively. Multiple-channel convolutions are used to independently learn the trajectories of different dimensions, improving the model's overall representations. 
$X_{i}^{H}$ includes the local correlated features of future trajectories in each history timestamp, and $X_{i}^{K}$ includes local correlated features of future trajectories in each mode. 
Align $X_{i}^{H}$ with $X_{i}^{K}$ by reshaping, then concatenate along the last dimension to form the multi-dimensional feature $X_{i}^{{}}$. 
An MLP with the Sigmoid activation function is used to decode $X_{i}$, 
generating the normalized value representing the relative position of anchor on the proposed trajectory.
The processes are denoted as:
\begin{equation}
	X_i = \text{Concat}(X_i^H, X_i^K)
\end{equation}
\begin{equation}
	\widehat{X_i} = \text{Sigmoid}(\text{MLP}(X_i))
\end{equation}
where $\widehat{X_i}$ is scaled by the future timestep range to select the trajectory point, which becomes the anchor ${{A}_{i}}\in {\mathbb{R}^{H\times K\times 2}}$.

%\subsubsection{Final Output}
\textbf{Final Output. }The proposed trajectories are converted to anchor-based polar coordinate representations, including relative distances and orientations. These features are encoded with MLP to obtain new mode queries.
Then, Agent-Map attention and Factorized attention are executed once more for the mode queries. 
Finally, the trajectory offset $\Delta P_{i}^{ref}$ and probability score $\pi _{i}^{{}}$ are produced by the MLP. 
The refined final trajectory is represented as:
\begin{equation}
	P_{i}^{fin} = P_{i}^{pro} + \Delta P_{i}^{{ref}}, \quad i \in [1, N]
\end{equation}
%-------------------------------------------------------------------------
\subsection{Training Loss}
The predictions of each timestamp include $\{P_{n,k}^{pro},P_{n,k}^{fin}\}$, ground truth as ${{g}_{n}}$. 
For multimodal trajectories, We take the winner-takes-all (WTA) \cite{lee2016stochastic} strategy, selecting the optimal $k_n$-th mode to calculate the loss. 
The selection for joint and marginal prediction tasks are, respectively, as follows:
\begin{equation}
	k_{n}^{jp} = \underset{k \in [1, K]}{\mathcal{\arg \min }}\, \sum\limits_{n=1}^{N} (P_{n,k} - g_{n}) 
\end{equation}
\begin{equation}
	k_{n}^{mp} = \underset{k \in [1, K]}{\mathcal{\arg \min }}\,(P_{n,k} - g_{n}) 
\end{equation}

The joint prediction task considers the minimum joint endpoint errors of all predicted trajectories within the same mode. 
Thus, $k_{{n}}^{jp}$ denotes the trajectory mode with the minimum joint endpoint error in a scenario, $k_{n}^{mp}$ denotes the trajectory mode with the minimum endpoint error for an agent.
The Huber loss function is used to calculate the regression loss for trajectory prediction:
\begin{equation} 
	\mathcal{L}_{reg}^{n}={\mathcal{L}_{Huber}}(P_{n,k_n}^{{}}-{{g}_{n}})
\end{equation}

The probabilistic classification loss for the final trajectory is obtained based on the cross-entropy loss function. 
Overall, the final loss of the model is defined as:
\begin{equation}
	\mathcal{L} = \frac{1}{TN} \sum\limits_{t=-F+1}^{0} \sum\limits_{n=1}^{N} (\mathcal{L}_{reg,pro}^{t,n} + \mathcal{L}_{reg,fin}^{t,n} + \mathcal{L}_{cls,fin}^{t,n}) 
\end{equation}
%-------------------------------------------------------------------------

\begin{figure*}[tp] % [t] 表示尽量放置在页面顶部
	\centering
	\includegraphics[width=0.98\textwidth]{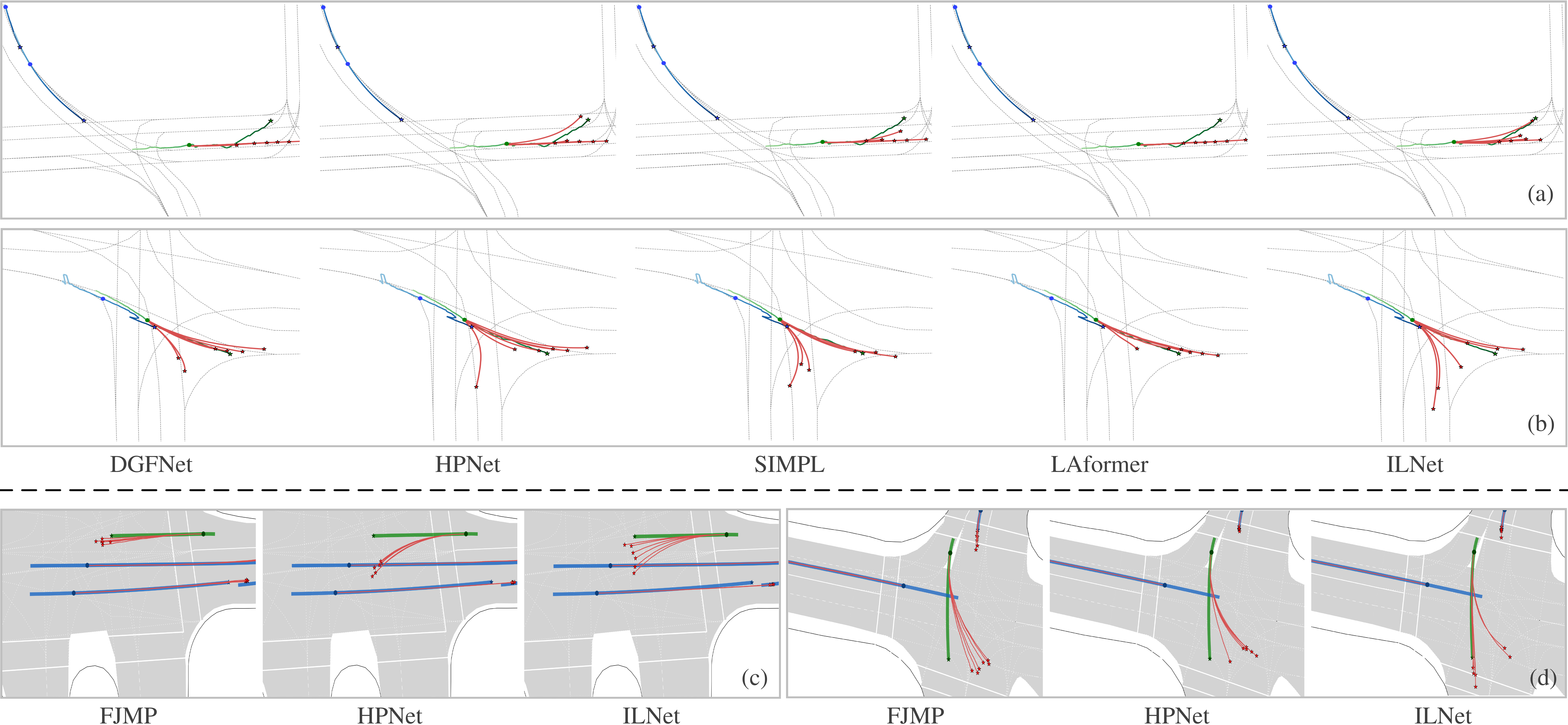} % 设置图片宽度为文本宽度的 100%
	
	\caption{Qualitative comparisons with recent works, (a) and (b) are in Argoverse, (c) and (d) are in INTERACTION. Predictions are indicated by red lines, each prediction is output at the final historical timestamp. Ground truth trajectories are indicated by green and blue lines.} % 设置图片标题
	\label{fig:fig7} % 设置图片标签，用于在文中引用
\end{figure*}

\section{Experiments}
\subsection{Experimental Settings}
\textbf{Datasets. }The performance of ILNet for joint multi-agent prediction on the INTERACTION \cite{zhan2019interaction} dataset and marginal prediction on the Argoverse \cite{chang2019argoverse} dataset is evaluated. 
INTERACTION dataset includes a variety of highly interactive scenarios, such as highway ramps, roundabouts, and intersections, with each trajectory comprising 1 second of past and 3 seconds of future data. 
Argoverse dataset consists of 300K scenarios, supporting 3-second future path prediction from 2 seconds of historical data. 
Both datasets offer high-definition maps with lane structures and are sampled at 10Hz. 
We follow the official dataset partitioning and valuate the test set results through the online leaderboards.

\textbf{Metrics. }We report the official metrics for joint prediction of INTERACTION: minimum Final Displacement Error between $K$ Joint future predictions from multiple agents and ground truth trajectories (minJointFDE),  minimum Average Displacement Error between $K$ Joint future predictions and ground truth trajectories (minJointADE). 
The official metrics for marginal prediction of Argoverse: minimum Average Displacement Error (minADE), minimum Final Displacement Error (minFDE), Miss Rate (MR) and Brier minimum Final Displacement Error (Brier-minFDE) of $K$ predictions for single agent. 
Brier-minFDE is calculated based on minFDE, summed with ${{(1-\pi )}^{2}}$, where $\pi$ denotes the probability score corresponding to the minFDE predictions. 
This metric is used to assess the joint accuracy of model's predicted optimal trajectory and its probability. 
For marginal and joint predictions, we uniformly set $k=6$.

To more fully evaluate the diversity quantity and quality in predicted trajectories, we also report the following metrics.

\textbullet\ Ratio of AvgFDE to MinFDE (RF) \cite{park2020diverse}: It measures the spread of $k$ predicted trajectory endpoints.

\textbullet\ Drivable Area Occupancy (DAO) \cite{park2020diverse}: It measures the proportion of pixels occupied by the $k$ predicted trajectory in the drivable area.

\textbullet\ Drivable Area Compliance (DAC): It measures the proportion of trajectory modes which doesn't go outside the drivable area.

\textbullet\ Average Angular Expansion (AAE) \cite{chen2024criteria}: Based on the vector formed by the start and end points of each trajectory, it calculates the pinch angle of vector pairs and then averages the results over all pairs.
The AAE aims to obtain the lateral diversity of trajectories.

\textbf{Implementation Details.}
Our model is trained end-to-end on 4 RTX 4090 GPUs for 64 epochs with a cosine annealing learning rate scheduler. The hidden dimension is 128.
In the trajectory refinement stage, we use the single layer for each multi-head attention with 8 heads. 
In the trajectory proposal stage, we apply an Inverse Learning attention layer and multiple Factorized attention layers.
Following the setup in HPNet \cite{tang2024hpnet}, we set the local area radius to 50 on Argoverse and 80 on INTERACTION. 
The initial learning rate is \( 3 \times 10^{-4} \) and \( 5 \times 10^{-4} \), respectively, both with a weight decay of \( 1 \times 10^{-4} \).
The Agent Future and Agent History attention search radius (\( R_f \) and \( R_h \)) are 100 for Argoverse and 80 for INTERACTION.
Due to hardware memory limitations, the batch size is 12 on Argoverse and 16 on INTERACTION. 
\vspace{-4mm}

\begin{table*}[th]
	\caption{Comparisons on the leaderboard of Argoverse motion Forecasting test set. There results are from single models. For each metric, the best result is in \textbf{bold} and the second best result is \underline{underlined}. B-minFDE$_{6}$ is the official ranking metric. Encoder represents the feature interaction method.}
	\centering
	\label{tab:tb2} 
	\centering
	\renewcommand{\arraystretch}{1.18} % 设置表格行距为1.6倍
	\scalebox{1.1}{ % 缩放表格到原始大小的90%
		\begin{tabular}{l|c|c|c|c|c|c|c}
			\hline
			\raisebox{-0.1\height}{Method} & \raisebox{-0.1\height}{Venue} & \raisebox{-0.1\height}{Encoder} & \raisebox{-0.1\height}{minFDE$_{6}\downarrow$} & \raisebox{-0.1\height}{MR$_{6}\downarrow$} & \raisebox{-0.1\height}{B-minFDE$_{6}\downarrow$} & \raisebox{-0.1\height}{minADE$_{6}\downarrow$} & \raisebox{-0.1\height}{Param(M)} \\ 
			\hline
			LaneGCN \cite{liang2020learning}           & ECCV 2020 & GCN & 1.362 & 0.162 & 2.054 & 0.870 & 3.7 \\
			mmTransformer \cite{liu2021multimodal}     & CVPR 2021 & Attention & 1.338 & 0.154 & 2.033 & 0.844 & 2.6 \\
			Scene Transformer \cite{ngiam2021scene}  & ICLR 2022 & Attention& 1.232 & 0.126 & 1.887 & 0.803 & 15.3 \\
			HiVT \cite{zhou2022hivt}              & CVPR 2022 & GAT & 1.169 & 0.127 & 1.842 & 0.774 & 2.5 \\
			GANet \cite{wang2023ganet}              & ICRA 2023 & GAT & 1.161 & 0.118 & 1.790 & 0.806 & 5.2 \\
			Macformer \cite{feng2023macformer}         & RAL 2023 & Attention & 1.216 & 0.121 & 1.827 & 0.819 & 2.4 \\ 
			STSCM \cite{liu2024reliable}			 & IF 2024 & GCN & 1.198 & \underline{0.107} & 1.792 & 0.804 & - \\
			SIMPL \cite{zhang2024simpl}             & RAL 2024 & Attention & 1.179 & 0.123 & 1.809 & 0.793 & 1.8 \\
			 HHLF \cite{jiao2024hierarchical}             & T-ITS 2024 & GAT + RL & 1.246 & 0.139 & - & 0.805 & -  \\
			LAformer \cite{liu2024laformer}       & CVPR 2024 & Attention & 1.163 & 0.125 & 1.835 & \underline{0.772} & 2.6 \\
			GoIRL \cite{peigoirl} & ICML 2025 & GAT + IRL & 1.173 & 0.120 & 1.796 & 0.809 & 3.6 \\
			DyMap\cite{fan2025bidirectional}	 & T-ASE 2025 & GAT & 1.129 & 0.110 & \textbf{1.729} & 0.795 & - \\
			DGFNet \cite{xin2025multi}            & RAL 2025 & 	Attention & \underline{1.117} & 0.108 & \underline{1.742} & \textbf{0.763} & 4.5 \\ 
			
			\rowcolor{gray!18} % 为最后一行添加灰色背景
			\textbf{ILNet} & Ours & GAT & \textbf{1.099} & \textbf{0.106}& 1.745& \underline{0.772} & 3.7    \\
			\hline
		\end{tabular}
	}
	
\end{table*}

\begin{table*}[h]
	\centering
	\caption{Comparisons on the leaderboard of INTERACTION multi-agent motion forecasting test set. For each metric, the best result is in \textbf{bold} and the second best result is \underline{underlined}.Encoder represents the feature interaction method.}
	\label{tab:tb1} 
	\renewcommand{\arraystretch}{1.18} % 设置表格行距
	\scalebox{1.1}{
		\begin{tabular}{l|c|c|c|c}
			\hline
			\raisebox{-0.1\height}{Method} & \raisebox{-0.1\height}{Venue} & \raisebox{-0.1\height}{Encoder} & \raisebox{-0.1\height}{minJointFDE$_{6}\downarrow$} & \raisebox{-0.1\height}{minJointADE$_{6}\downarrow$} \\ 
			\hline
			DenseTNT \cite{gu2021densetnt} & ICCV 2021 & Attention& 1.130  & 0.420 \\ 
			AutoBot \cite{girgis2022latent} & ICLR 2022 & Attention & 1.015 & 0.312 \\ 
			THOMAS \cite{gilles2022thomas} & ICLR 2022 & GCN & 0.968 & 0.416 \\ 
			HDGT \cite{jia2023hdgt} & T-PAMI 2023 & GAT& 0.958 & 0.303 \\ 
			Traj-MAE \cite{chen2023traj} & CVPR 2023 & Attention& 0.966 & 0.307 \\ 
			FJMP \cite{rowe2023fjmp} & CVPR 2023 & GAT& 0.922 & 0.275 \\  
			HPNet \cite{tang2024hpnet} & CVPR 2024 & GAT& \underline{0.823} & 0.255 \\
			GAMDTP \cite{liu2025gamdtp} & Arxiv 2025 &  GAT + Mamba & 0.830 & \underline{0.253}\\
			HyperMTP \cite{lu2025hyper} & IF 2025 &  GAT & 0.937 & 0.296\\

			\rowcolor{gray!18} % 灰色背景
			\textbf{ILNet} & Ours & GAT & \textbf{0.777} & \textbf{0.248} \\  
			\hline
		\end{tabular}
	}
\end{table*}
\vspace{-2mm}
\subsection{Comparison with State-of-the-art Methods}
\textbf{Comparison of quantitative results. }The marginal trajectory prediction accuracy of ILNet is evaluated on the Agroverse, and the single-model results are shown in Table. \ref{tab:tb2}. 
The GAT is Graph Attention Network, the GCN is Graph Convolutional Network, the RL is Reinforcement Learning, and the IRL is Inverse Reinforcement Learning.
Our ILNet achieves state-of-the-art performance of the metrics minFDE and MR with fewer parameters, and is highly competitive in other metrics. 
Compared to LanGCN with the same number of parameters, ILNet achieves a 19\% reduction in minFDE. 
For DyMap with the best Brier-minFDE, our method has significant advantages in the remaining metrics.
Additionally, ILNet reduces minFDE by 0.018 compared to DGFNet, using 0.8M fewer parameters. 

The joint trajectory prediction results of our ILNet on the INTERACTION multi-agent track are shown in Table. \ref{tab:tb1}. 
On the test set, our method outperforms state-of-the-art methods and ranks first place. 
Compared to the HPNet, ILNet achieves a decrease of 2.7\% in minJointADE and 5.6\% in minJointFDE. 
Moreover, the parameters of ILNet are 18.9\% smaller than HPNet (4.3M versus 5.3M). 
We attribute the significant enhancement of results to the proposed Inverse Learning attention, which effectively improves the model's ability to capture the complex interaction intentions, as can be seen later in the qualitative results.

\textbf{Comparison of qualitative results. }We select the scenarios with higher uncertainty in the Argoverse and INTERACTION validation sets, making the fair visual comparisons between ILNet and recent works which provide the official checkpoint files, as shown in Fig. \ref{fig:fig7}. 
Notably, the FJMP utilizes the validation set for training. 
In Fig. \ref{fig:fig7} (a), both DGFNet and LAformer demonstrate limitations in predicting turning modals. 
Compared with HPNet and SIMPL, our method has higher accuracy and more modals are similar to the ground truth.
In Fig. \ref{fig:fig7} (b), our ILNet balances accuracy while predicting three distinctly different modals, it achieves the best trajectory diversity.
In Fig. \ref{fig:fig7} (c), our ILNet predicts both straight-ahead and diverse turning modals. In contrast, FJMP has more homogeneous trajectories, and HPNet fails to predict the modal that matches the ground truth.
In Fig. \ref{fig:fig7} (d), our method captures not only the obvious trajectory modals, but also deeper modals which are the same as the ground truth.
Such impressive results underscore the effectiveness of the proposed ILNet. 
In contrast to HPNet, which forward models the temporal consistency between successively predicted trajectories, ILNet first utilizes the visible future state to perform guided inverse  reasoning about past interactions. It transforms prediction into a more sound logical inference.
This enables ILNet to capture more subtle intentions by understanding the reasons behind operations, thus providing more comprehensive and accurate historical predictions.
Moreover, it can provide greater flexibility for downstream tasks such as decision planning.

\begin{table}[h]
	\centering
	\caption{Ablation studies of IL attention and DAS module. Experiments are performed on INTERACTION validation set.}
	\renewcommand{\arraystretch}{1.4} % 设置表格行距为1.1倍
	\scalebox{1.03}{
		\begin{tabular}{c|cccc|c|c@{\hspace{4pt}}c@{\hspace{2pt}}} % 在最左侧添加一列
			\hline
			\multirow{2}{*}{\raisebox{-0.2\height}{ID}} & \multicolumn{4}{c|}{\raisebox{-0.1\height}{IL Attention}} & \multirow{2}{*}{\raisebox{-0.2\height}{DAS}} & \multicolumn{2}{c}{\raisebox{-0.1\height}{Metrics}} \\ \cline{2-5} \cline{7-8}
			& \raisebox{-0.3\height}{TA} & \raisebox{-0.3\height}{FA} & \raisebox{-0.3\height}{HA} & \raisebox{-0.3\height}{IL} &  & \raisebox{-0.2\height}{mJointFDE$_{6}\downarrow$} & \raisebox{-0.2\height}{mJointADE$_{6}\downarrow$} \\ 
			\hline
			1 & \checkmark &           &           &           &           & 0.562 & 0.179 \\ 
			2 & \checkmark &\checkmark &           &           &           & 0.554 & 0.173 \\ 
			3 & \checkmark &           &\checkmark &           &           & 0.554 & 0.175 \\ 
			4 & \checkmark &\checkmark &\checkmark &           &           & 0.551 & 0.172 \\ 
			5 & \checkmark &\checkmark &\checkmark &  \checkmark         &           & 0.540 & 0.170 \\ 
			6 & \checkmark &           &           & & \checkmark & 0.557 & 0.177 \\ 
			%\rowcolor{gray!18} % 为最后一行添加灰色背景 
			7 & \checkmark &\checkmark &\checkmark &\checkmark &\checkmark & \textbf{0.537} & \textbf{0.167} \\ 
			\hline
		\end{tabular}
	}
	\label{tab:tb3} 
\end{table}

\begin{table*}[h]
	\centering
	\renewcommand{\arraystretch}{1.4}
	\caption{Ablation studies of FA + HA module and DAS module for different scenario types on the INTERACTION validation set.}
	\label{tab:joint_metrics}

	\scalebox{1}{
		\begin{tabular}{l|cc|cc|cc}
			\hline
			\multirow{2}{*}{\raisebox{-0.2\height}{Method}} & 
			\multicolumn{2}{c|}{Intersection} & 
			\multicolumn{2}{c|}{Merging} & 
			\multicolumn{2}{c}{Roundabout} \\ \cline{2-7}
			& \raisebox{-0.1\height}{minJointADE$_{6}$} & \raisebox{-0.1\height}{minJointFDE$_{6}$} 
			& \raisebox{-0.1\height}{minJointADE$_{6}$} & \raisebox{-0.1\height}{minJointFDE$_{6}$} 
			& \raisebox{-0.1\height}{minJointADE$_{6}$} & \raisebox{-0.1\height}{minJointFDE$_{6}$} \\
			\hline
			w/o FA+HA + w/o DAS & 0.192 & 0.612 & 0.114 & 0.347 & 0.203 & 0.642 \\
			w/ FA+HA + w/o DAS  & 0.180 & 0.586 & 0.113 & 0.342 & 0.191 & 0.612 \\
			w/ FA+HA + w/ DAS   & \textbf{0.177} & \textbf{0.574} & \textbf{0.112} & \textbf{0.336} & \textbf{0.189} & \textbf{0.606} \\
			\hline
		\end{tabular}
}
\end{table*}

\subsection{Analysis and Discussion}
\textbf{Ablation study. }We conduct ablation studies on the IL attention and DAS module to verify the effectiveness of each design, results summarized in Table. \ref{tab:tb3}. 
The Ego-Agent Temporal attention, Agent Future attention, and Agent History attention in ILNet are represented by "TA", "FA", and "HA", respectively, while "IL" denotes the inverse learning modeling.
The baseline configuration (ID-1) retains only the Ego-Agent Temporal attention.
The performance improvements of the single FA (ID-3) and single HA (ID-4) module are evaluated separately, proving the correctness of modeling interactions across spatial and temporal scales. 
A comparative analysis between the forward paradigm (HA+FA, ID-5) and the inverse paradigm (FA+HA, ID-6) for encoding proposed intentions reveals that the inverse encoding strategy brings significant performance earnings over the forward approach.
ID-6 decreases minJointFDE by 3.9\% and minJointADE by 5\% compared to ID-1.
This phenomenon validates the benefits of introducing inverse learning attention to capture interaction behavior, underscoring bidirectional modeling provides a more comprehensive perspective.
Furthermore, introducing the DAS module (ID-7) achieves a decrease of 0.9\% in minJointFDE and 1.1\% in minJointADE. 
The full pipeline, integrating the FA+HA+DAS modules (ID-8), achieves the best performance, demonstrating a 4.4\% reduction in minJointFDE and a 6.7\% reduction in minJointADE.

We also present a special ablation study for three typical scenarios with different interaction characteristics, to deeply evaluate the synergistic effectiveness of our proposed module in different traffic scenarios, results are summarized in Table. \ref{tab:joint_metrics}.
The FA+HA module demonstrates significant improvements over the DAS module in highly complex interactive scenarios such as intersections and roundabouts, underscoring its critical role in capturing sophisticated interactive intentions. Notably, in merging scenarios (dense traffic but simpler interactions) the DAS module's contribution to reducing the final endpoint error surpassed that of the FA+HA module. 
This suggests that the strategy of dynamically identifying key trajectory anchor is universally applicable, with its advantages undiminished in less interactive settings. 
In summary, these findings reveal a powerful synergy between the two modules: the former endows the model with a profound understanding of interactions, while the latter performs universal trajectory refinement upon this foundation.

\begin{table*}[tp]
	\centering
	\caption{Quantitative comparisons with latest state-of-the-art methods. The best result is in \textbf{bold} and the second best result is \underline{underlined}. Baseline* is the variant without the FA+HA but with the DAS module.}
	\renewcommand{\arraystretch}{1.44} % 设置表格行距
	\scalebox{1.03}{
		\begin{tabular}{l|cccc|cccc|c|c}
			\hline
			\multirow{2}{*}{\raisebox{-0.3\height}{Method}} & \multicolumn{4}{c|}{Accuracy} & \multicolumn{4}{c|}{Diversity} & \multirow{2}{*}{\raisebox{-0.3\height}{Infer Time(ms)$\downarrow$}} & \multirow{2}{*}{\raisebox{-0.3\height}{Param(M)$\downarrow$}} \\ \cline{2-9} 
			& \raisebox{-0.1\height}{B-minFDE$_{6}\downarrow$} & \raisebox{-0.1\height}{minADE$_{6}\downarrow$} & \raisebox{-0.1\height}{minFDE$_{6}\downarrow$} & \raisebox{-0.1\height}{MR$_{6}\downarrow$} & 
			\raisebox{-0.1\height}{AAE$\uparrow$} & 
			\raisebox{-0.1\height}{DAO$\uparrow$} & 
			\raisebox{-0.1\height}{DAC$\uparrow$} & 
			\raisebox{-0.1\height}{RF$\uparrow$} & & \\ 
			\hline
			DGFNet\cite{xin2025multi} & 2.306 & 0.991 & 1.665 & 0.235 & 7.652 & \textbf{59.784} & 0.982 & 3.724 & \textbf{30.7} & 4.5\\ 
			HPNet\cite{tang2024hpnet} & \underline{2.288} & \underline{0.972} & 1.611 & \textbf{0.182} & 9.687 & 57.886 & \textbf{0.988} & 3.851 & 39.5 & 4.1\\ 				\hline
			Baseline* & 2.291 & 0.981 & \underline{1.596} & 0.196 & \underline{9.725} & 58.447 & \underline{0.986} & \underline{3.886} & \underline{38.3} & \textbf{3.2}\\ 
			ILNet & \textbf{2.209} & \textbf{0.948} & \textbf{1.531} & \underline{0.184} & \textbf{9.945} & \underline{58.982} & \textbf{0.988} & \textbf{4.082} & 41.2 & \underline{3.7}\\ 				
			\hline
		\end{tabular}
	}
	\label{tab:tb4}
\end{table*}

\begin{figure*}[h] % [h] 表示尽量在当前位置插入图片
	\centering
	\includegraphics[width=0.95\textwidth]{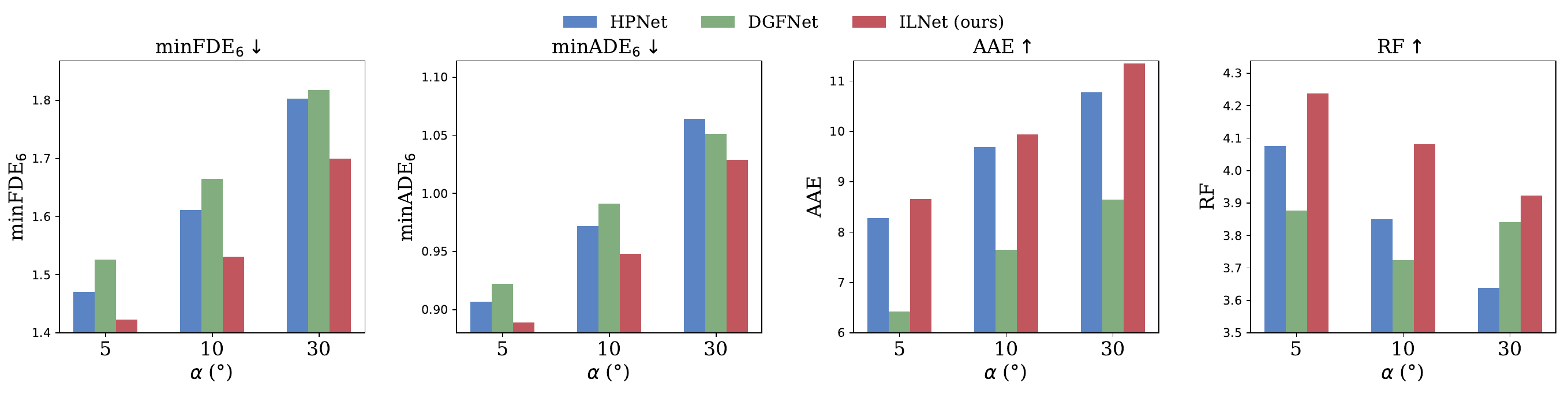} % 设置图片宽度为文本宽度的 90% 和高度为文本高度的 50%
	\caption{Comparisons of predictive accuracy (minFDE$_{6}\downarrow$, minADE$_{6}\downarrow$) and diversity (AAE$\uparrow$, RF$\uparrow$) with latest state-of-the-art methods for scenarios which satisfy absolute angle $\alpha$ on the Argoverse validation set.} % 设置图片标题
	\label{fig:fig8} % 设置图片标签，用于在文中引用
\end{figure*}

\begin{figure*}[h] % [h] 表示尽量在当前位置插入图片
	\centering
	\includegraphics[width=1\textwidth, height=0.24\textheight]{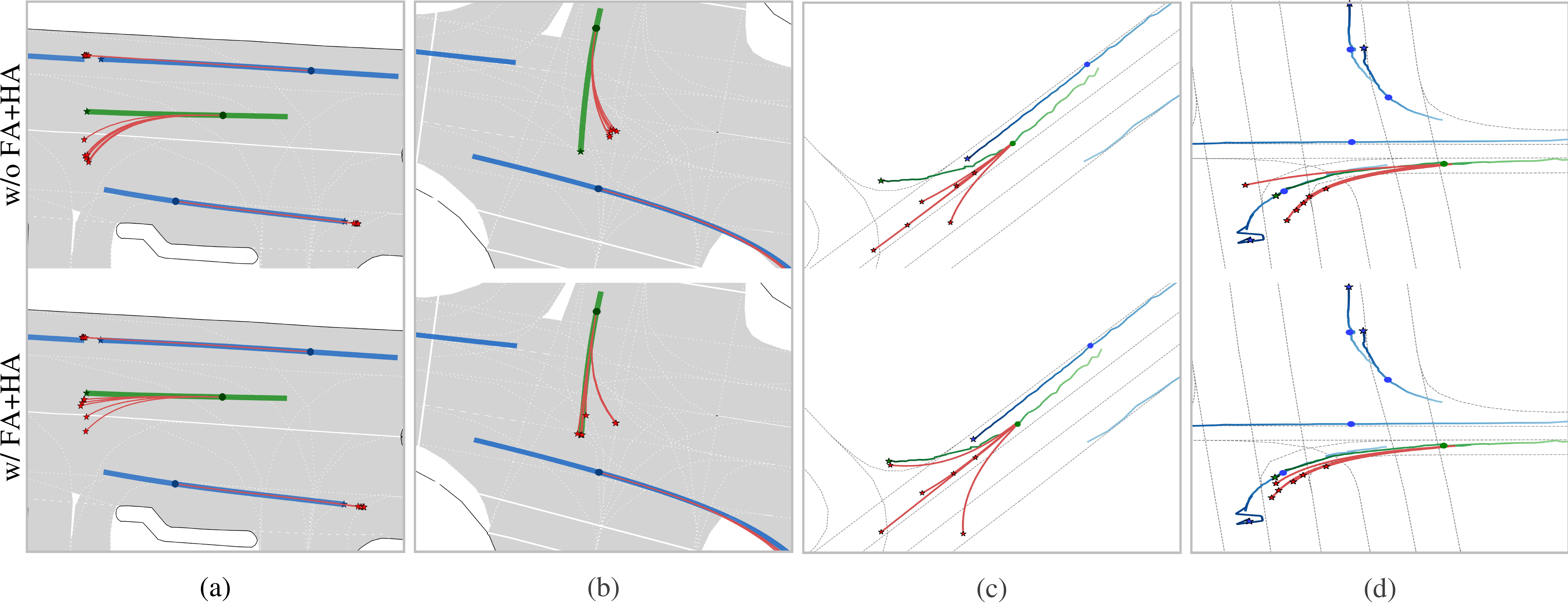} % 设置图片宽度为文本宽度的 90% 和高度为文本高度的 50%
	\caption{Qualitative results on different validation sets, (a) and (b) are in INTERACTION, (c) and (d) are in Argoverse. Predictions are indicated by red lines, each prediction is output at the final historical timestamp. Ground truth trajectories are indicated by green and blue lines.} % 设置图片标题
	\label{fig:fig5} % 设置图片标签，用于在文中引用
\end{figure*}

\textbf{Performance comparison on highly interactive scenarios. }We are specifically  interested in the predictive accuracy and diversity of ILNet in complex interactive scenarios, as it's critical to developing robust and reliable autonomous systems.
However, the agent trajectories in the Argoverse dataset are usually regular straight-ahead following or turning. 
Therefore, we select the scenarios which simultaneously satisfy the following four conditions to be regarded as challenging scenarios in Argoverse: the road intersection scenario \cite{chen2024criteria}; 
the scenario with the focal agent FDE more than $d=5$ meters based on constant velocity modeling \cite{rowe2023fjmp}; 
the scenario in which at least one agent is within 5 meters of the focal agent and the interaction time is not less than 25 timestamps;
the scenario in which the absolute angle $\alpha$ of at least 10° between the future and historical ground truth endpoints of the focal agent.
We obtain 429 challenging scenarios from the Argoverse validation set, and conduct quantitative comparisons of accuracy and diversity between ILNet and the latest state-of-the-art methods.
All methods are evaluated using official checkpoint files under identical inference conditions to ensure fair comparison. 
The results are shown in Table. \ref{tab:tb4}. 
ILNet achieves substantial improvements across all metrics compared to the Baseline* without the FA+HA, with the inference latency increasing by only an average of three ms. 
This highlights its efficiency in great performance gain with little latency overhead.
Compared to the latest state-of-the-art methods, we achieve the best performance on all metrics except MR and DAO with fewer parameters, evidencing better predictive accuracy and diversity.
Additionally, ILNet maintains competitive inference latency to ensure fast and scalable in real-time applications.
We also present the comparison of the prediction performance between different methods under the increasing absolute angle of focal agent, as shown in Fig. \ref{fig:fig8}.
As the absolute angle $\alpha$ increases, the prediction becomes more challenging. However, our method exhibits greater generalization and effectiveness compared to other methods, as evidenced by the widening performance gap.

\textbf{Importances of FA+HA. }We qualitatively compare the predictive performance with and without the FA+HA model on different datasets, as shown in Fig. \ref{fig:fig5} (a-d).
It can be seen that the introduction of FA+HA has better prediction results. 
In Fig. \ref{fig:fig5} (a), The historical trajectory of green agent displays a possible left turn across the opposite lane and interaction with the blue agent in that lane. 
The w/o FA+HA model predicts only the turning modes, which are the direct intentions. 
However, the w/ FA+HA model more accurately captures the intention.
Its predicted trajectories not only remain distinct turning modes, but also have straight-ahead modes, which are more consistent with the ground truth.
In addition, the w/ FA+HA model exhibits better trajectory diversity due to its deep extraction of interaction intention.
In Fig. \ref{fig:fig5} (b), the green agent enters the intersection, in which case its intention is more difficult to determine because future behavior can be either straight-ahead or left turn.
The w/o FA+HA model consistently predicts all modes as left turns, whereas the w/ FA+HA model accurately predicts four straight-ahead modes that align with the true values, along with two left turn modes.
The above shows that FA+HA can significantly improve the intention extraction ability of model thereby delivering more reliable predictions, especially in complex scenarios. 
In Fig. \ref{fig:fig5} (c), the two agents are traveling in the same lane, and the blue agent is located at the right rear of the green agent. 
It is not difficult to predict the driving intention in the lane. 
However, a mode query in the w/ FA+HA model is accurately decoded as the trajectory to out of lane, consistent with the ground truth. 
This demonstrates the ability of FA+HA to extract more subtle intentions, further validating its effectiveness.
Fig. \ref{fig:fig5} (d) shows the following interaction scenario. 
The prediction results show that the w/o FA+HA model can't accurately extract further intentions of following, exhibiting separation of trajectories.
On the other hand, the w/ FA+HA model is more effective in capturing latent intention, therefore has better prediction accuracy and stability.

\begin{table}[h]
	\centering
	\renewcommand{\arraystretch}{1.4} % 设置表格行距为1.18倍
	\caption{Results of different IL Attention search radii $ R_f $ and $ R_h $.}
	
	\scalebox{0.93}{
		\begin{tabular}{c|ccc|cc}
			\hline
			\multirow{2}{*}{\raisebox{-0.2\height}{\( R_f,_h \)}} & \multicolumn{3}{c|}{Argoverse} & \multicolumn{2}{c}{INTERACTION} \\ \cline{2-6} 
			& \raisebox{-0.2\height}{minFDE$_{6}$} & \raisebox{-0.2\height}{minADE$_{6}$} & \raisebox{-0.2\height}{MR$_{6}$} & \raisebox{-0.2\height}{minJointFDE$_{6}$} & \raisebox{-0.2\height}{minJointADE$_{6}$}  \\
			\hline
			1x   & 0.869 & 0.645 & 0.071 & \textbf{0.537} & \textbf{0.167} \\
			1.5x & 0.870 & 0.645 & 0.070 & 0.537 & 0.167 \\
			2x   & \textbf{0.867} & \textbf{0.644} & \textbf{0.070} & 0.538 & 0.168 \\
			\hline
		\end{tabular}
	}
	\label{tab:tb6} 
\end{table}

\textbf{Effects of IL Attention search radius.} We study the sensitivity of model performance to the IL Attention search radius ($ R_f $ and $ R_h $).
The radius is set as the local area radius (1x), where INTERACTION is 80 and Argoverse is 50. 
As well as the radius is expanded by 1.5x and 2x respectively. 
The results are summarized in Table. \ref{tab:tb6}.
For INTERACTION, the default local area radius yields the best performance, suggesting that a more focused receptive field is sufficient to capture the dense and short-range interactions.
For Argoverse, the best performance is achieved with the 2x scaled radius (i.e., 100), highlighting the need for a broader receptive field to capture equally critical long-range interactions in open urban environments, thereby enabling more robust predictions.

\begin{table}[h]
		\centering
		\renewcommand{\arraystretch}{1.4}
		\caption{Results of different agent history trajectory random mask Ratios on the Argoverse validation set.}
		
		\scalebox{0.95}{
		\begin{tabular}{c|cc|cc}
			\hline
			\multirow{2}{*}{\raisebox{-0.2\height}{Traj Mask Ratio (\%)}} & \multicolumn{2}{c|}{\raisebox{-0.1\height}{w/ FA+HA}} & \multicolumn{2}{c}{\raisebox{-0.1\height}{w/o FA+HA}} \\ \cline{2-5} 
			& \raisebox{-0.2\height}{minFDE$_{6}$} & \raisebox{-0.2\height}{minADE$_{6}$} & \raisebox{-0.2\height}{minFDE$_{6}$} & \raisebox{-0.2\height}{minADE$_{6}$} \\
			\hline
			10 & 2.75  & 1.49 & 2.84  & 1.54 \\
			30 & 6.66  & 3.26 & 6.81  & 3.37 \\
			50 & 10.50 & 5.03 & 10.75 & 5.18 \\
			\hline
		\end{tabular}
	}
		\label{tab:tb7} 
\end{table}

\begin{figure}[h] % [h] 表示尽量在当前位置插入图片
	\centering
	\includegraphics[width=0.46\textwidth]{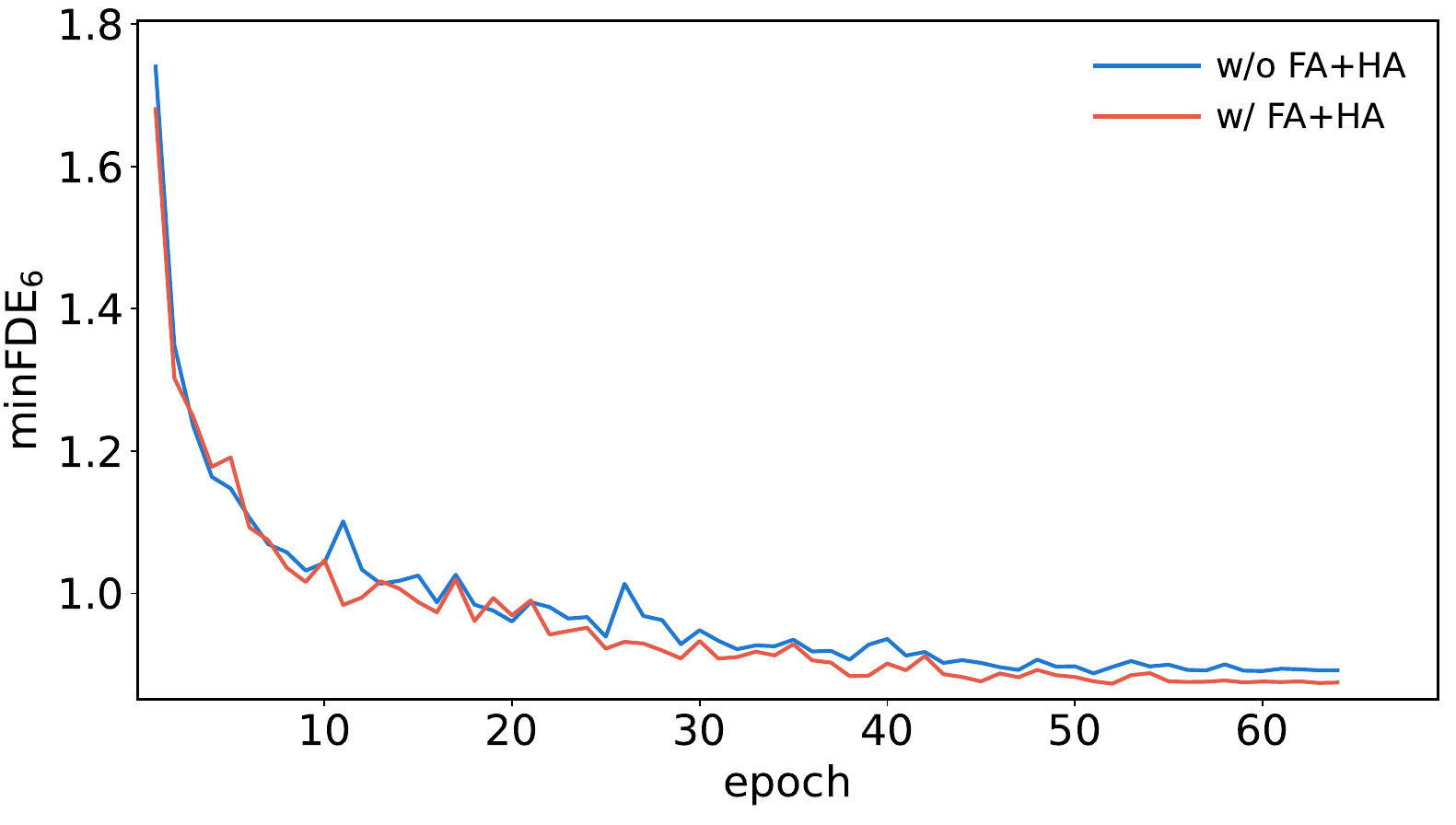} % 设置图片宽度为文本宽度的 90% 和高度为文本高度的 50%
	\caption{Performance curves on the Argoverse validation set, evaluated at the end of each training epoch.} % 设置图片标题
	\label{fig:fig9} % 设置图片标签，用于在文中引用
	\vspace{-1em} % 调整间距
\end{figure}

\textbf{Stability analysis of FA+HA. }To evaluate the stability of FA+HA against incomplete observations, we apply different random mask ratios to the history trajectory of all agents in the scenario.
The results in Table. \ref{tab:tb7}.
Even with limited or incomplete data, our w/ FA+HA method approach significantly outperforms the w/o FA+HA approach at all levels. 
This sustained advantage suggests that by explicitly reasoning about the relationship between past and future states, our approach can better reconstruct and infer plausible intentions from degraded historical contexts. 
This highlights the superior resilience of the inverse learning mechanism.
We also visualize the Performance curve on the Argoverse validation set, as shown in Fig. \ref{fig:fig9}.
The w/ FA+HA method (red curve) not only converges to a lower final error but also demonstrates a markedly faster convergence rate compared to the w/o FA+HA method (blue curve).
Unlike forward modeling which requires complex, open-ended extrapolation, our inverse modeling reconstructs the problem into a more constrained task of inferring the logical link between adjacent past and future states. 
It provides the model with a more direct and comprehensible learning objective from the outset, reducing initial uncertainty and guiding it towards a more optimal solution path more rapidly. 

\begin{figure}[h] % [h] 表示尽量在当前位置插入图片
	\centering
	\includegraphics[width=0.47\textwidth]{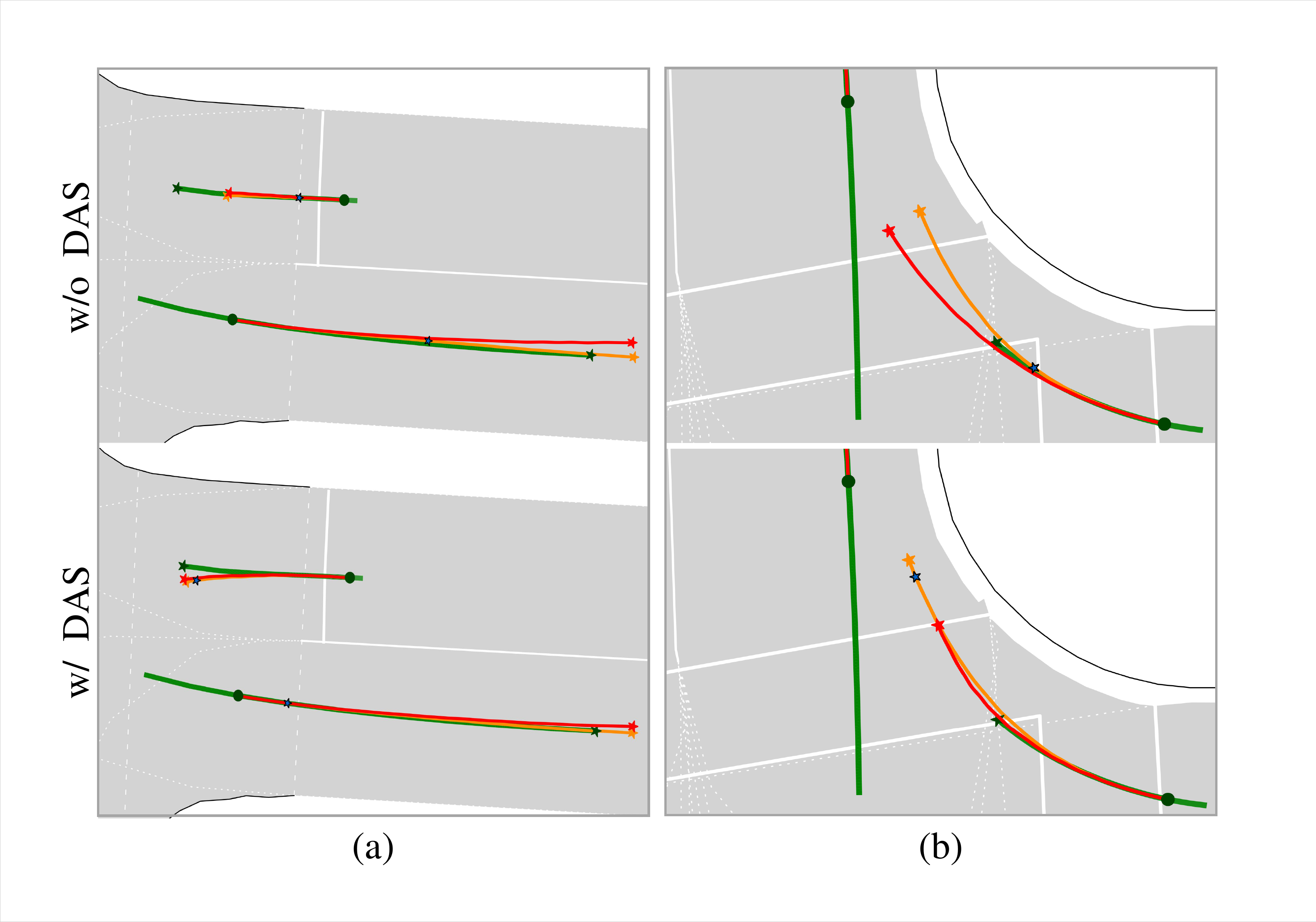} % 设置图片宽度为文本宽度的 70%
	\caption{Qualitative results on the INTERACTION validation set. The yellow and red lines are the proposed and final trajectories respectively, the blue stars indicate the location of anchors in proposed trajectories. } % 设置图片标题
	\label{fig:fig6} % 设置图片标签，用于在文中引用
\end{figure}

\textbf{Importances of DAS. }Fig. \ref{fig:fig6} illustrates some qualitative results with and without the DAS model on the INTERACTION validation set.
Note that each prediction used for comparison is decoded based on the same $k$-th mode query, the w/o DAS model uses a fixed trajectory midpoint strategy, 
following HPNet. 
In Fig. \ref{fig:fig6} (a), the left agent's proposed trajectory (yellow line) is more consistent with the ground truth. 
However, in the refinement stage, the w/o DAS model uses the midpoint of proposed trajectory as the anchor (blue star), and the refined trajectory (red line) deviates from the ground truth. 
In contrast, the w/ DAS model selects a position near the end of history trajectory as the anchor, and has less trajectory offset. 
Another result in Fig. \ref{fig:fig6} (b), where a right agent turns right at the intersection, the w/o DAS model refines the proposed trajectory and reduces error. 
The w/ DAS model selects the end of proposed trajectory as the anchor, capturing more future interactions, and the refinement result is closer to ground truth. 
These results underscore the effectiveness of dynamic anchor selection strategy, which empowers the model to focus on more important future contexts, leading to more stable and accurate final predictions.

\begin{table}[ht]
	\centering
	\renewcommand{\arraystretch}{1.6} % 设置表格行距为1.18倍
	\small
	\caption{Model performance and inference time evaluated on the validation set.}
	\scalebox{0.85}{
		\begin{tabular}{c|c|c|c|c|c|c}
			\hline
			& Method & Param(M) & T & Inf(ms) & mJointFDE$\downarrow$ & mJointADE$\downarrow$ \\ \cline{2-7} 
			&                                                  &                         & s              & 37.6                   &                         &                         \\ \cline{4-5}
			& \multirow{-2}{*}{w/o DAS}                        & \multirow{-2}{*}{4.3}  & p              & 91.4                   & \multirow{-2}{*}{0.540} & \multirow{-2}{*}{0.170}  \\ \cline{2-7} 
			&                          &                         & s              & 38.4                   &                         &                         \\ 
			\cline{4-5}
			\multirow{-5}{*}{\begin{minipage}[c]{0.2cm}\begin{center}\rotatebox{90}{INTERACTION}\end{center}\end{minipage}} & \multirow{-2}{*}{w/ DAS} & \multirow{-2}{*}{+0.004} & p              & 93.5                  & 
			\multirow{-2}{*}{0.537} & \multirow{-2}{*}{0.167} \\ \hline
			& Method & Param(M) & T & Inf(ms) & mFDE$_{6}\downarrow$ & b\_mFDE$_{6}\downarrow$ \\ \cline{2-7} 
			&                                                  &                         & s              & 40.3                   &                         &                         \\ \cline{4-5}
			& \multirow{-2}{*}{w/o DAS}                        & \multirow{-2}{*}{3.7}  & p              & 141.8                  & \multirow{-2}{*}{0.875} & \multirow{-2}{*}{1.526} \\ \cline{2-7} 
			&                          &                         & s              & 41.2                   &                         &                         \\ 
			
			\cline{4-5}
			\multirow{-5}{*}{\begin{minipage}[c]{0.2cm}\begin{center}\rotatebox{90}{Argoverse}\end{center}\end{minipage}} & \multirow{-2}{*}{w/ DAS} & \multirow{-2}{*}{+0.012} & p              & 144.4                  & \multirow{-2}{*}{0.867} & \multirow{-2}{*}{1.519} \\ \hline
		\end{tabular}
	}
	
	\label{tab:tb5} 
\end{table}
\textbf{Inference time of DAS. }The multi-agent inference time of the DAS module on Argovere and INTERACTION validation sets is measured by using the same GPU, the results are shown in Table. \ref{tab:tb5}.  
The inference processing tracks include serial (s) and parallel (p), with the number of parallel inference scenarios being 16 and 12 for INTERACTION and Argovere, respectively.
It can be seen that the introduction of the DAS module barely increases the number of parameters in the whole model. 
With only a minimal increase in serial and parallel inference time, the performance on both datasets is noticeably improved, which clearly indicates the importance of successive predictions to be refined by considering dynamic anchors.

\textbf{Accuracy-efficiency analysis of different DAS strategies}.
To explicitly investigate the accuracy-latency trade-off within the DAS module, we conducted a study comparing our proposed design with both a lighter variant (w/o Conv) and a more complex one (using two anchors to refine different parts of each proposed trajectory ). 
The results in Table. \ref{tab:my_label}.
Removing the 2D convolutional layers (DAS (w/o Conv)) slightly reduces latency but leads to a noticeable drop in accuracy, demonstrating that the CNNs are crucial for effective feature extraction. 
Conversely, increasing the complexity by using two anchors (DAS (2 Anchors)) offers a negligible improvement in final error (a mere 0.001 reduction in minJointFDE) at the cost of an 8ms increase in latency, due to the repeated refinement process. 
This analysis confirms that our proposed single-anchor DAS module occupies the sweet spot in the trade-off spectrum, providing the best balance of high accuracy and computational efficiency.
\begin{table}[h]
	\centering
	\renewcommand{\arraystretch}{1.4}
	\caption{Results of different DAS strategies on the INTERACTION validation set.}
	\label{tab:my_label}
	
	\scalebox{0.95}{
		\begin{tabular}{c|c|c|c}
			\hline
			Method & minJointADE$_{6}$ & minJointFDE$_{6}$ & Infer Time (ms) \\
			\hline
			DAS (w/o Conv)              & 0.170                & 0.539                & 38.0                    \\
			DAS               & 0.167                & 0.537                & 38.4                    \\
			DAS (2 Anchors)   & 0.167                & 0.536                & 46.2                    \\
			\hline
		\end{tabular}
}
\end{table}
\vspace{-1em} % 调整间距
\section{Conclusion}
This paper introduces an ILNet framework for multi-agent trajectory prediction. 
A novel Inverse Learning (IL) attention is designed to inversely capture the intentions of agent interactions, based on the visible future states in their history. 
Our framework aims to dynamically encode the spatio-temporal coordination of historical trajectory interactions, which enhances the model's ability to understand delicate intentions. 
Besides, we propose a Dynamic Anchor Selection module for extracting trajectory change keypoints in parallel. It improves performance with nearly no increase in the number of parameters.
Extensive experimental results show that ILNet surpasses methods with fewer parameters on the INTERACTION and Argoverse datasets. 
Specifically in challenged interaction scenarios, ILNet achieves higher accuracy and more multimodal distributions of trajectories over fewer parameters.
The future work will focus on three main directions: 
1) integrating generative models and diversity loss functions into the proposal stage, in order to achieve a wider and more accurate set of initial predictions.
2) Investigating methods for decomposing the latent space of pattern queries. The goal is for each trajectory output by the model to be associated with a specific, interpretable manipulation.
3) training a more lightweight model via knowledge distillation and introducing an adaptive search radius for the Inverse Learning attention to dynamically adjust computational cost based on scene complexity. Furthermore, a stream-based feature encoding mode can be adopted during deployment to minimize redundant processing by reusing historical computation results, thereby significantly enhancing the model's real-time performance.


\begin{thebibliography}{}
\providecommand{\url}[1]{#1}
\csname url@samestyle\endcsname
\providecommand{\newblock}{\relax}
\providecommand{\bibinfo}[2]{#2}
\providecommand{\BIBentrySTDinterwordspacing}{\spaceskip=0pt\relax}
\providecommand{\BIBentryALTinterwordstretchfactor}{4}
\providecommand{\BIBentryALTinterwordspacing}{\spaceskip=\fontdimen2\font plus
\BIBentryALTinterwordstretchfactor\fontdimen3\font minus
  \fontdimen4\font\relax}
\providecommand{\BIBforeignlanguage}[2]{{%
\expandafter\ifx\csname l@#1\endcsname\relax
\typeout{** WARNING: IEEEtran.bst: No hyphenation pattern has been}%
\typeout{** loaded for the language `#1'. Using the pattern for}%
\typeout{** the default language instead.}%
\else
\language=\csname l@#1\endcsname
\fi
#2}}
\providecommand{\BIBdecl}{\relax}
\BIBdecl

\end{thebibliography}


\begin{thebibliography}{1}
\bibliographystyle{IEEEtran}

\bibitem{liang2020learning}
M. Liang, B. Yang, R. Hu, Y. Chen, R. Liao, S. Feng, and R. Urtasun, ``Learning lane graph representations for motion forecasting,'' in \textit{Proc. Eur. Conf. Comput. Vis. (ECCV)}, Glasgow, UK, 2020, pp. 541--556.

\bibitem{mozaffari2020deep}
S. Mozaffari, O. Y. Al-Jarrah, M. Dianati, P. Jennings, and A. Mouzakitis, ``Deep learning-based vehicle behavior prediction for autonomous driving applications: A review,'' \textit{IEEE Trans. Intell. Transp. Syst.}, vol. 23, no. 1, pp. 33--47, 2020.

\bibitem{deng2025social}
Z. Deng, W. Hu, T. Huang, C. Sun, J. Zhong, and A. Khajepour, 
``Social predictive intelligent driver model for autonomous driving simulation,'' 
\textit{Automot. Innov.}, vol. None, pp. 1--12, 2025.

\bibitem{phan2020covernet}
T. Phan-Minh, E. C. Grigore, F. A. Boulton, O. Beijbom, and E. M. Wolff, ``Covernet: Multimodal behavior prediction using trajectory sets,'' in \textit{Proc. IEEE/CVF Conf. Comput. Vis. Pattern Recognit. (CVPR)}, 2020, pp. 14074--14083.

\bibitem{chai2019multipath}
Y. Chai, B. Sapp, M. Bansal, and D. Anguelov, ``Multipath: Multiple probabilistic anchor trajectory hypotheses for behavior prediction,'' \textit{arXiv preprint arXiv:1910.05449}, 2019.

\bibitem{hong2019rules}
J. Hong, B. Sapp, and J. Philbin, ``Rules of the road: Predicting driving behavior with a convolutional model of semantic interactions,'' in \textit{Proc. IEEE/CVF Conf. Comput. Vis. Pattern Recognit. (CVPR)}, 2019, pp. 8454--8462.

\bibitem{wang2023ganet}
M. Wang, X. Zhu, C. Yu, W. Li, Y. Ma, R. Jin, X. Ren, D. Ren, M. Wang, and W. Yang, ``Ganet: Goal area network for motion forecasting,'' in \textit{Proc. IEEE Int. Conf. Robot. Autom. (ICRA)}, 2023, pp. 1609--1615.

\bibitem{zeng2021lanercnn}
W. Zeng, M. Liang, R. Liao, and R. Urtasun, ``Lanercnn: Distributed representations for graph-centric motion forecasting,'' in \textit{Proc. IEEE/RSJ Int. Conf. Intell. Robots Syst. (IROS)}, 2021, pp. 532--539.

\bibitem{gao2020vectornet}
J. Gao, C. Sun, H. Zhao, Y. Shen, D. Anguelov, C. Li, and C. Schmid, ``Vectornet: Encoding hd maps and agent dynamics from vectorized representation,'' in \textit{Proc. IEEE/CVF Conf. Comput. Vis. Pattern Recognit. (CVPR)}, 2020, pp. 11525--11533.

\bibitem{ye2021tpcn}
M. Ye, T. Cao, and Q. Chen, ``Tpcn: Temporal point cloud networks for motion forecasting,'' in \textit{Proc. IEEE/CVF Conf. Comput. Vis. Pattern Recognit. (CVPR)}, 2021, pp. 11318--11327.

\bibitem{nikhil2018convolutional}
N. Nikhil and B. Tran Morris, ``Convolutional neural network for trajectory prediction,'' in \textit{Proc. Eur. Conf. Comput. Vis. (ECCV) Workshops}, 2018, pp. 0--0.

\bibitem{mohamed2020social}
A. Mohamed, K. Qian, M. Elhoseiny, and C. Claudel, ``Social-stgcnn: A social spatio-temporal graph convolutional neural network for human trajectory prediction,'' in \textit{Proc. IEEE/CVF Conf. Computer Vision Pattern Recognition}, 2020, pp. 14424--14432.

\bibitem{zhang2019sr}
P. Zhang, W. Ouyang, P. Zhang, J. Xue, and N. Zheng, ``SR-LSTM: State refinement for LSTM towards pedestrian trajectory prediction,'' in \textit{Proc. IEEE/CVF Conf. Computer Vision Pattern Recognition}, 2019, pp. 12085--12094.

\bibitem{yu2020spatio}
C. Yu, X. Ma, J. Ren, H. Zhao, and S. Yi, ``Spatio-temporal graph transformer networks for pedestrian trajectory prediction,'' in \textit{Proc. Computer Vision--ECCV 2020: 16th European Conf.}, Glasgow, UK, Aug. 2020, pp. 507--523.

\bibitem{zhao2021tnt}
H. Zhao, J. Gao, T. Lan, C. Sun, B. Sapp, B. Varadarajan, Y. Shen, Y. Shen, Y. Chai, C. Schmid, and others, ``TNT: Target-driven trajectory prediction,'' in \textit{Proc. Conf. Robot. Learn.}, 2021, pp. 895--904.

\bibitem{gu2021densetnt}
J. Gu, C. Sun, and H. Zhao, ``Densetnt: End-to-end trajectory prediction from dense goal sets,'' in \textit{Proc. IEEE/CVF Int. Conf. Computer Vision}, 2021, pp. 15303--15312.

\bibitem{gilles2022gohome}
T. Gilles, S. Sabatini, D. Tsishkou, B. Stanciulescu, and F. Moutarde, ``GoHome: Graph-oriented heatmap output for future motion estimation,'' in \textit{Proc. 2022 Int. Conf. Robotics and Automation (ICRA)}, 2022, pp. 9107--9114.

\bibitem{nayakanti2023wayformer}
N. Nayakanti, R. Al-Rfou, A. Zhou, K. Goel, K. S. Refaat, and B. Sapp, ``Wayformer: Motion forecasting via simple \& efficient attention networks,'' in \textit{Proc. 2023 IEEE Int. Conf. Robotics and Automation (ICRA)}, 2023, pp. 2980--2987.

\bibitem{jiao2024hierarchical}
Y. Jiao, M. Miao, Z. Yin, C. Lei, X. Zhu, X. Zhao, L. Nie, and B. Tao, 
``A hierarchical hybrid learning framework for multi-agent trajectory prediction,'' 
\textit{IEEE Trans. Intell. Transp. Syst.}, vol. 25, no. 8, pp. 10344--10354, 2024.

\bibitem{ngiam2021scene}
J. Ngiam, B. Caine, V. Vasudevan, Z. Zhang, H.-T. L. Chiang, J. Ling, R. Roelofs, A. Bewley, C. Liu, A. Venugopal, and others, ``Scene transformer: A unified architecture for predicting multiple agent trajectories,'' \textit{arXiv preprint arXiv:2106.08417}, 2021.

\bibitem{zhou2023query}
Z. Zhou, J. Wang, Y.-H. Li, and Y.-K. Huang, ``Query-centric trajectory prediction,'' in \textit{Proc. IEEE/CVF Conf. Computer Vision Pattern Recognition}, 2023, pp. 17863--17873.

\bibitem{tang2024hpnet}
X. Tang, M. Kan, S. Shan, Z. Ji, J. Bai, and X. Chen, ``HPNet: Dynamic trajectory forecasting with historical prediction attention,'' in \textit{Proc. IEEE/CVF Conf. Computer Vision Pattern Recognition}, 2024, pp. 15261--15270.

\bibitem{zhou2023qcnext}
Z. Zhou, Z. Wen, J. Wang, Y.-H. Li, and Y.-K. Huang, ``QCNext: A next-generation framework for joint multi-agent trajectory prediction,'' \textit{arXiv preprint arXiv:2306.10508}, 2023.

\bibitem{liao2025diffusiondrive}
B. Liao, S. Chen, H. Yin, B. Jiang, C. Wang, S. Yan, X. Zhang, X. Li, Y. Zhang, Q. Zhang, et al., 
``DiffusionDrive: Truncated diffusion model for end-to-end autonomous driving,'' in 
\textit{Proc. IEEE Conf. Comput. Vis. Pattern Recognit.}, 2025, pp. 12037--12047.

\bibitem{deng2024eliminating}
Z. Deng, W. Hu, C. Sun, D. Chu, T. Huang, W. Li, C. Yu, M. Pirani, D. Cao, A. Khajepour, and Y. Chao, 
``Eliminating uncertainty of driver’s social preferences for lane change decision-making in realistic simulation environment,'' 
\textit{IEEE Trans. Intell. Transp. Syst.}, 2024.

\bibitem{lu2025hyper}
Y. Lu, W. Wang, R. Bai, S. Zhou, L. Garg, A. K. Bashir, W. Jiang, and X. Hu, 
``Hyper-relational Interaction Modeling in Multi-modal Trajectory Prediction for Intelligent Connected Vehicles in Smart Cities,'' 
\textit{Inf. Fusion}, vol. 114, pp. 102682, 2025.

\bibitem{sun2018probabilistic}
L. Sun, W. Zhan, and M. Tomizuka, ``Probabilistic prediction of interactive driving behavior via hierarchical inverse reinforcement learning,'' in \textit{Proc. 2018 21st Int. Conf. Intelligent Transportation Systems (ITSC)}, 2018, pp. 2111--2117.

\bibitem{shi2024mtr++}
S. Shi, L. Jiang, D. Dai, and B. Schiele, ``MTR++: Multi-agent motion prediction with symmetric scene modeling and guided intention querying,'' \textit{IEEE Trans. Pattern Anal. Machine Intell.}, 2024.

\bibitem{liu2025gamdtp}
Y. Liu, H. Niu, and J. Zhu, 
``GAMDTP: Dynamic trajectory prediction with graph attention Mamba network,'' 
\textit{arXiv:2504.04862}, 2025.

\bibitem{wang2023prophnet}
X. Wang, T. Su, F. Da, and X. Yang, ``ProphNet: Efficient agent-centric motion forecasting with anchor-informed proposals,'' in \textit{Proc. IEEE/CVF Conf. Computer Vision Pattern Recognition}, 2023, pp. 21995--22003.

\bibitem{zhou2024smartrefine}
Y. Zhou, H. Shao, L. Wang, S. L. Waslander, H. Li, and Y. Liu, ``SmartRefine: A scenario-adaptive refinement framework for efficient motion prediction,'' in \textit{Proc. IEEE/CVF Conf. Computer Vision Pattern Recognition}, 2024, pp. 15281--15290.

\bibitem{chen2024criteria}
C. Chen, M. Pourkeshavarz, and A. Rasouli, ``Criteria: A new benchmarking paradigm for evaluating trajectory prediction models for autonomous driving,'' in \textit{Proc. IEEE Int. Conf. Robotics Autom. (ICRA)}, 2024, pp. 8265--8271.

\bibitem{park2020diverse}
S. H. Park, G. Lee, J. Seo, M. Bhat, M. Kang, J. Francis, A. Jadhav, P. P. Liang, and L. P. Morency, ``Diverse and admissible trajectory forecasting through multimodal context understanding,'' in \textit{Proc. Eur. Conf. Comput. Vis. (ECCV)}, Glasgow, UK, 2020, vol. 16, pp. 282--298.

\bibitem{liu2024reliable}
J. Liu, H. Lin, X. Wang, L. Wu, S. Garg, and M. M. Hassan, 
``Reliable trajectory prediction in scene fusion based on spatio-temporal structure causal model,'' 
\textit{Inf. Fusion}, vol. 107, p. 102309, 2024.

\bibitem{wissing2018trajectory}
C. Wissing, T. Nattermann, K.-H. Glander, and T. Bertram, ``Trajectory prediction for safety critical maneuvers in automated highway driving,'' in \textit{Proc. 2018 21st Int. Conf. Intelligent Transportation Systems (ITSC)}, 2018, pp. 131--136.

\bibitem{schlechtriemen2015will}
J. Schlechtriemen, F. Wirthmueller, A. Wedel, G. Breuel, and K.-D. Kuhnert, ``When will it change the lane? A probabilistic regression approach for rarely occurring events,'' in \textit{Proc. 2015 IEEE Intelligent Vehicles Symp. (IV)}, 2015, pp. 1373--1379.

\bibitem{li2020social}
J. Li, H. Ma, Z. Zhang, and M. Tomizuka, ``Social-WAGDAT: Interaction-aware trajectory prediction via Wasserstein graph double-attention network,'' \textit{arXiv preprint arXiv:2002.06241}, 2020.

\bibitem{liu2024diftraj}
Y. Liu, X. Dong, Y. Lin, and M. Ye, 
``DifTraj: Diffusion Inspired by Intrinsic Intention and Extrinsic Interaction for Multi-Modal Trajectory Prediction,'' in 
\textit{Proc. Int. Joint Conf. Artif. Intell.}, 2024.


\bibitem{alahi2016social}
A. Alahi, K. Goel, V. Ramanathan, A. Robicquet, L. Fei-Fei, and S. Savarese, ``Social LSTM: Human trajectory prediction in crowded spaces,'' in \textit{Proc. IEEE Conf. Computer Vision and Pattern Recognition}, 2016, pp. 961--971.

\bibitem{liu2021multimodal}
Y. Liu, J. Zhang, L. Fang, Q. Jiang, and B. Zhou, ``Multimodal motion prediction with stacked transformers,'' in \textit{Proc. IEEE/CVF Conf. Computer Vision Pattern Recognition}, 2021, pp. 7577--7586.

\bibitem{hu2022scenario}
Y. Hu, W. Zhan, and M. Tomizuka, ``Scenario-transferable semantic graph reasoning for interaction-aware probabilistic prediction,'' \textit{IEEE Trans. Intell. Transp. Syst.}, vol. 23, no. 12, pp. 23212--23230, 2022.

\bibitem{chen2022intention}
X. Chen, H. Zhang, F. Zhao, Y. Hu, C. Tan, and J. Yang, ``Intention-aware vehicle trajectory prediction based on spatial-temporal dynamic attention network for internet of vehicles,'' \textit{IEEE Trans. Intell. Transp. Syst.}, vol. 23, no. 10, pp. 19471--19483, 2022.

\bibitem{li2021grin}
L. Li, J. Yao, W. Li, T. He, T. Xiao, J. Yan, D. Wipf, and Z. Zhang, ``GRIN: Generative relation and intention network for multi-agent trajectory prediction,'' \textit{Adv. Neural Inf. Process. Syst.}, vol. 34, pp. 27107--27118, 2021.

\bibitem{mo2022multi}
X. Mo, Z. Huang, Y. Xing, and C. Lv, ``Multi-agent trajectory prediction with heterogeneous edge-enhanced graph attention network,'' \textit{IEEE Trans. Intell. Transp. Syst.}, vol. 23, no. 7, pp. 9554--9567, 2022.

\bibitem{li2020evolvegraph}
J. Li, F. Yang, M. Tomizuka, and C. Choi, ``Evolvegraph: Multi-agent trajectory prediction with dynamic relational reasoning,'' \textit{Adv. Neural Inf. Process. Syst.}, vol. 33, pp. 19783--19794, 2020.

\bibitem{xin2025multi}
G. Xin, D. Chu, L. Lu, Z. Deng, Y. Lu, and X. Wu, ``Multi-agent trajectory prediction with difficulty-guided feature enhancement network,'' \textit{IEEE Robotics and Automation Letters}, 2025.

\bibitem{liu2024laformer}
M. Liu, H. Cheng, L. Chen, H. Broszio, J. Li, R. Zhao, M. Sester, and M. Y. Yang, ``Laformer: Trajectory prediction for autonomous driving with lane-aware scene constraints,'' in \textit{Proc. IEEE/CVF Conf. Computer Vision Pattern Recognition}, 2024, pp. 2039--2049.

%\bibitem{wu2023tsgn}
%Y. Wu, T. Gilles, B. Stanciulescu, and F. Moutarde, ``TSGN: Temporal Scene Graph Neural Networks with Projected Vectorized Representation for Multi-Agent Motion Prediction,'' in \textit{Proc. 2023 IEEE Intell. Vehicles Symp. (IV)}, 2023, pp. 1--8.

\bibitem{peigoirl}
M. Pei, S. Shi, L. Zhang, P. Li, and S. Shen, 
``GoIRL: Graph-Oriented Inverse Reinforcement Learning for Multimodal Trajectory Prediction,'' in \textit{Proc. Forty-second International Conference on Machine Learning}, 2025.

\bibitem{salzmann2020trajectron++}
T. Salzmann, B. Ivanovic, P. Chakravarty, and M. Pavone, ``Trajectron++: Dynamically-feasible trajectory forecasting with heterogeneous data,'' in \textit{Computer Vision--ECCV 2020: 16th European Conf., Glasgow, UK, Aug. 23--28, 2020, Proc. Part XVIII}, 2020, pp. 683--700.

\bibitem{yuan2021agentformer}
Y. Yuan, X. Weng, Y. Ou, and K. M. Kitani, ``Agentformer: Agent-aware transformers for socio-temporal multi-agent forecasting,'' in \textit{Proc. IEEE/CVF Int. Conf. Computer Vision}, 2021, pp. 9813--9823.

\bibitem{zhou2022hivt}
Z. Zhou, L. Ye, J. Wang, K. Wu, and K. Lu, ``HIVT: Hierarchical vector transformer for multi-agent motion prediction,'' in \textit{Proc. IEEE/CVF Conf. Computer Vision Pattern Recognition}, 2022, pp. 8823--8833.

\bibitem{fan2025bidirectional}
B. Fan, H. Yuan, Y. Dong, Z. Zhu, and H. Liu, ``Bidirectional agent-map interaction feature learning leveraged by map-related tasks for trajectory prediction in autonomous driving,'' \textit{IEEE Transactions on Automation Science and Engineering}, vol. 2025, doi: 10.1109/TASE.2025.3529736.

\bibitem{sadeghian2019sophie}
A. Sadeghian, V. Kosaraju, A. Sadeghian, N. Hirose, H. Rezatofighi, and S. Savarese, ``SOPHIE: An attentive GAN for predicting paths compliant to social and physical constraints,'' in \textit{Proc. IEEE/CVF Conf. Computer Vision Pattern Recognition}, 2019, pp. 1349--1358.

%\bibitem{kang2024ffinet}
%M. Kang, S. Wang, S. Zhou, K. Ye, J. Jiang, and N. Zheng, ``FFINet: Future feedback interaction network for motion forecasting,'' \textit{IEEE Transactions on Intelligent Transportation Systems}, vol. 25, no. 9, pp. 12285--12296, 2024, doi: 10.1109/TITS.2024.3381631.

\bibitem{varadarajan2022multipath++}
B. Varadarajan, A. Hefny, A. Srivastava, K. S. Refaat, N. Nayakanti, A. Cornman, K. Chen, B. Douillard, C. P. Lam, D. Anguelov, and others, ``Multipath++: Efficient information fusion and trajectory aggregation for behavior prediction,'' in \textit{Proc. 2022 Int. Conf. Robotics and Automation (ICRA)}, 2022, pp. 7814--7821.

\bibitem{zhang2022trajectory}
L. Zhang, P. Li, J. Chen, and S. Shaojie, ``Trajectory prediction with graph-based dual-scale context fusion,'' in \textit{Proc. 2022 IEEE/RSJ Int. Conf. Intelligent Robots and Systems (IROS)}, 2022, pp. 11374--11381.

\bibitem{choi2023r}
S. Choi, J. Kim, J. Yun, and J. W. Choi, ``R-pred: Two-stage motion prediction via tube-query attention-based trajectory refinement,'' in \textit{Proc. IEEE/CVF Int. Conf. Computer Vision}, 2023, pp. 8525--8535.

\bibitem{shi2022motion}
S. Shi, L. Jiang, D. Dai, and B. Schiele, ``Motion transformer with global intention localization and local movement refinement,'' \textit{Adv. Neural Inf. Process. Syst.}, vol. 35, pp. 6531--6543, 2022.

\bibitem{ye2023bootstrap}
M. Ye, J. Xu, X. Xu, T. Wang, T. Cao, and Q. Chen, ``Bootstrap motion forecasting with self-consistent constraints,'' in \textit{Proc. IEEE/CVF Int. Conf. Computer Vision}, 2023, pp. 8504--8514.

\bibitem{feng2023macformer}
C. Feng, H. Zhou, H. Lin, Z. Zhang, Z. Xu, C. Zhang, B. Zhou, and S. Shaojie, ``Macformer: Map-agent coupled transformer for real-time and robust trajectory prediction,'' \textit{IEEE Robotics and Automation Letters}, 2023.

\bibitem{zhang2024simpl}
L. Zhang, P. Li, S. Liu, and S. Shaojie, ``SIMPL: A Simple and Efficient Multi-agent Motion Prediction Baseline for Autonomous Driving,'' \textit{IEEE Robotics and Automation Letters}, 2024.

\bibitem{rowe2023fjmp}
L. Rowe, M. Ethier, E.-H. Dykhne, and K. Czarnecki, ``FJMP: Factorized joint multi-agent motion prediction over learned directed acyclic interaction graphs,'' in \textit{Proc. IEEE/CVF Conf. Computer Vision Pattern Recognition}, 2023, pp. 13745--13755.

\bibitem{chen2023traj}
H. Chen, J. Wang, K. Shao, F. Liu, J. Hao, C. Guan, G. Chen, and P.-A. Heng, ``Traj-MAE: Masked autoencoders for trajectory prediction,'' in \textit{Proc. IEEE/CVF Int. Conf. Computer Vision}, 2023, pp. 8351--8362.

\bibitem{jia2023hdgt}
X. Jia, P. Wu, L. Chen, Y. Liu, H. Li, and J. Yan, ``HDGT: Heterogeneous driving graph transformer for multi-agent trajectory prediction via scene encoding,'' \textit{IEEE Trans. Pattern Anal. Mach. Intell.}, 2023.

\bibitem{gilles2022thomas}
T. Gilles, S. Sabatini, D. Tsishkou, B. Stanciulescu, and F. Moutarde, ``THOMAS: Trajectory heatmap output with learned multi-agent sampling,'' in \textit{Proc. Int. Conf. Learn. Represent. (ICLR)}, 2022.

\bibitem{girgis2022latent}
R. Girgis, F. Golemo, F. Codevilla, M. Weiss, J. A. D'Souza, S. E. Kahou, F. Heide, and C. Pal, ``Latent variable sequential set transformers for joint multi-agent motion prediction,'' in \textit{Proc. Int. Conf. Learn. Represent. (ICLR)}, 2022.

\bibitem{lee2016stochastic}
S. Lee, S. P. Purushwalkam, M. Cogswell, V. Ranjan, D. Crandall, and D. Batra, ``Stochastic multiple choice learning for training diverse deep ensembles,'' \textit{Adv. Neural Inf. Process. Syst.}, vol. 29, 2016.

\bibitem{chang2019argoverse}
M.-F. Chang, J. Lambert, P. Sangkloy, J. Singh, S. Bak, A. Hartnett, D. Wang, P. Carr, S. Lucey, D. Ramanan, and others, ``Argoverse: 3D tracking and forecasting with rich maps,'' in \textit{Proc. IEEE/CVF Conf. Computer Vision Pattern Recognition}, 2019, pp. 8748--8757.

\bibitem{zhan2019interaction}
W. Zhan, L. Sun, D. Wang, H. Shi, A. Clausse, M. Naumann, J. Kummerle, H. Konigshof, C. Stiller, A. de La Fortelle, and others, ``Interaction dataset: An international, adversarial and cooperative motion dataset in interactive driving scenarios with semantic maps,'' \textit{arXiv preprint arXiv:1910.03088}, 2019.


\end{thebibliography}
\end{document}